\documentclass[sigplan,10pt,final,screen,nonacm]{acmart}
\setcopyright{none}
\usepackage{amsmath}
\usepackage{amsfonts}
\usepackage{graphicx}
\usepackage{booktabs}
\usepackage{color}
\usepackage{algpseudocode}
\usepackage{listings}
\usepackage{multirow}
\usepackage{pgfplots}
\usepackage{pgf-pie}
\usepackage{todonotes}
\usepackage{url}
\usepackage{subcaption}
\usepackage{tikz}
\usetikzlibrary{positioning}
\usetikzlibrary{calc}
\pdfoutput=1 

\newcommand{\fb}[1]{\dofb#1}
\newcommand{\dofb}[1]{\textbf{#1}\nobreak\hspace{0pt}}

\makeatletter
\g@addto@macro{\UrlBreaks}{\UrlOrds}
\let\OldStatex\Statex
\renewcommand{\Statex}[1][3]{%
  \setlength\@tempdima{\algorithmicindent}%
  \OldStatex\hskip\dimexpr#1\@tempdima\relax}
\makeatother

\pgfplotsset{compat=1.15}
\begin{document}

%\special{papersize=8.5in,11in}
%\setlength{\pdfpageheight}{\paperheight}
%\setlength{\pdfpagewidth}{\paperwidth}

%\conferenceinfo{CONF 'yy}{Month d--d, 20yy, City, ST, Country} 
%\copyrightyear{20yy} 
%\copyrightdata{978-1-nnnn-nnnn-n/yy/mm} 
%\doi{nnnnnnn.nnnnnnn}
\acmConference[CGO'20]{Code Generation and Optimization}

\title{Learning Branch Probabilities in Compiler from Datacenter Workloads}

\author{Easwaran Raman}
\affiliation{Google}
\email{eraman@google.com}

\author{Xinliang David Li}
\affiliation{Google}
\email{davidxl@google.com}

\begin{abstract}
Estimating the probability with which a conditional branch instruction is taken is an important analysis that enables many optimizations in modern compilers. When using Profile Guided Optimizations (PGO), compilers are able to make a good estimation of the branch probabilities. In the absence of profile information, compilers resort to using heuristics for this purpose. In this work, we propose learning branch probabilities from a large corpus of data obtained from datacenter workloads. 

Using metrics including \emph{Root Mean Squared Error}, \emph{Mean Absolute Error} and \emph{cross-entropy}, we show that the machine learning model improves branch probability estimation by 18-50\% in comparison to compiler heuristics. This translates to performance improvement of up to 8.1\% on 24 out of a suite of 40 benchmarks with a 1\% geomean improvement on the suite. This also results in greater than 1.2\% performance improvement in an important search application.

\end{abstract}

\maketitle

\begin{abstract}
\input{abs}
\end{abstract}

%\category{CR-number}{subcategory}{third-level}

% general terms are not compulsory anymore, 
% you may leave them out
%\terms
%term1, term2

%\keywords
%keyword1, keyword2
%\input{outline}
\section{Introduction}
\label{sec:intro}
Profile Guided Optimizations (PGO) \cite{hubicka:2005,li:2010} is a powerful optimization technique that is available in modern optimizing compilers. In PGO mode, compilers use execution profiles that are obtained either from running an instrumented version of the binary or from hardware performance counters to guide optimizations. The execution profile consumed by the LLVM compiler consists of the following:
\begin{itemize}
    \item For each control transfer instruction such as a branch or a switch instruction, the probabilities of the control getting transferred to each of its targets.
    \item The execution count or the sample count at the entry point of each function in the application.
    \item The histogram of values used by particular instructions such as indirect function call targets.
    \item A concise global summary of the execution counts.
\end{itemize}
For the purpose of this work, we focus only on the target probabilities of control transfer instructions. We further limit ourselves to two-way conditional branch instructions. We use the term \emph{branch probability} to refer to the probability with which two-way conditional branch instruction is taken. This information is used by many optimizations among which code layout is the most significant. 

In the absence of profile information, compilers use heuristics \cite{wu:1994} to estimate the branch probabilities. These heuristics apply only to a small set of branches. For instance, one heuristic treats the back-edge of a loop as a highly taken branch. For a large fraction of branches, compilers simply assume that the branch is unbiased (i.e. both branch targets are equally likely). However, in a large collection of more than 7 million branches, we observe that only 43\% can be actually considered unbiased.

In this work, we present a machine learning~\cite{murphy:2012} system to improve the estimation of branch probabilities without profile information. This machine learning (ML) system learns branch behavior from  production workloads running on datacenters. We use Google Wide Profiling~\cite{ren:2010} to collect Last Branch Record~\cite{intelmanual} or LBR data from processors in the datacenter. This is converted into SamplePGO~\cite{chen:2016} profile format. The workloads are then compiled in SamplePGO mode in LLVM using these profiles to generate training data for the ML system. For each branch, we collect a set of features and its branch probability derived from profile data. The branch probabilities obtained from the profile serve as the labels for training a {\em supervised machine learning model}. The trained model can then be used for compiling any code to improve the accuracy of branch probability estimation without requiring profiles.

The contributions of this work are
\begin{itemize}
\item We present the first ML system to estimate branch \emph{probabilities} in a C++ compiler that significantly outperforms the heuristics used in the compiler.

\item Our ML system learns from  real production workloads running on  datacenters. This results in a training data of millions of examples, which is orders of magnitude larger than what is used by similar works in the past.

\item We demonstrate that the improved branch probability estimation results in better performance of an important datacenter workload and a large benchmark suite.

\end{itemize}

The rest of this paper is organized as follows. Section \ref{sec:related} discusses related work in the area of machine learning use in compilers. Section~\ref{sec:motivation} describes the utility of branch probability estimation in the compiler and why machine learning is a suitable approach to estimate branch probabilities. The use of machine learning to infer branch probabilities is described in detail in Section \ref{sec:training}. Section \ref{sec:results} evaluates the machine learning approach and Section \ref{sec:conclusion} concludes the paper.
\section{Related Work}
\label{sec:related}
The work most closely related to this work is Corpus Based Static Prediction by Calder et al.~\cite{Calder:1995} and \cite{calder:1996}. Their work also trains a neural network on features of static branch instructions. The crucial differences are:
\begin{itemize}
    \item Our system learns branch \emph{probabilities} of LLVM IR-level branch instructions. The probability information learned by the system is used to optimize the code in the compiler. In their work, they use the model to predict the \emph{direction} of static branches from a trace of machine instructions and that information is not used by compiler optimizations.
    \item The system we propose learns from real workload running on a datacenter. This results in a much larger training set and also allows the model to be continuously updated as the behavior of workload changes. In contrast, the training data for their system is obtained from a set of 40 benchmarks.
    \item We use a deeper neural network with more layers to improve the accuracy.
\end{itemize}

Jim{\'e}nez et al~\cite{jimenez:2002, jimenez:2003} have proposed using neural network methods to predict the direction of \emph{dynamic} branch instructions. Prediction of dynamic branch instructions in hardware constrains the machine learning techniques that can be used. 

Wagner et al.~\cite{wagner:1994} go beyond simple heuristics to estimate the relative frequency of basic blocks, a problem closely related to branch  probability estimation. In their work, they model the control flow within a function as a Markov process.

There have been many efforts in using machine learning and related techniques to replace hand-crafted heuristics in compilers. Wang and O'Boyle~\cite{wang:2018} present an extensive survey of the use of machine learning in compiler optimizations. Stephenson and Amarasinghe~\cite{stephenson:2005} used supervised learning to predict loop unrolling factors. Their training data consisted of around 2500 examples, which is much smaller than what we use in this work. Eliot et al.~\cite{eliot:nips1997} used supervised learning technique to train a local (single basic block) instruction scheduler. It uses a machine model to predict the so called preference relationship given a partial schedule and two candidate/ready instructions. Simon et al.~\cite{simon:2013} used unsupervised learning to automatically construct inline heuristics.

Instruction scheduling is a hard problem in the compiler that extensively uses heuristics. The application of learning to instruction scheduling within straight line code has been explored by Moss et al.~\cite{moss:1998} and McGovern et al.~\cite{mcgovern:2002}. 

The optimization that benefits the most with better branch probability estimation is code layout. Code layout results in performance improvement by making \emph{instructions} flow into the processor pipeline without stalls. Data prefetching plays a similar role in bringing \emph{data} into the processor without stalls. 
Traditionally, compiler assisted data prefetching insertion is based on two strategies: heuristic based insertion for accesses with linear stride, and stride profiling based approach~\cite{wu:stride}. In ~\cite{milad:icml2018}, Hashemi et al. treated the memory prefetching strategies as an n-gram classification problem in natural language processing, and used LSTM based Recurrent Nerual Network (RNN) to do the prediction. Peled et al.~\cite{peled:2015} defines the notion of \emph{semantic locality} and use reinforcement learning techniques to build a context-based memory prefetcher that approximates semantic locality. 

The works listed above apply learning techniques to specific optimizations. In contrast Cummins et al.~\cite{cummins:pact2017} used Deep Learning to learn optimization heuristics automatically.
Another approach to optimize programs without dealing with specific optimizations is \emph{super-optimization}. This refers to the process of finding a better version of a given program that is semantically equivalent. Early efforts in super-optimization relied on brute force search. Recent efforts have focused on using stochastic search to improve the efficiency. Bunel et al.~\cite{bunel:2016} have used reinforcement learning to optimize stochastic search based super-optimization techniques.

\section{Background and Motivation}
\label{sec:motivation}

%\begin{figure*}
%    \centering
%    \begin{subfigure}[t]{0.3}
%        \centering
%        \includegraphics[scale=0.2]{CFG.png}
%        \label{fig:layout_example_a}
%    \end{subfigure}
%    \begin{subfigure}[t]{0.3}
%        \centering
%        \includegraphics[scale=0.2]{Layout.png}
%        \label{fig:layout_example_b}
%    \end{subfigure}
%    \caption{Code Layout}
%    \label{fig:layout_example}
%\end{figure*}

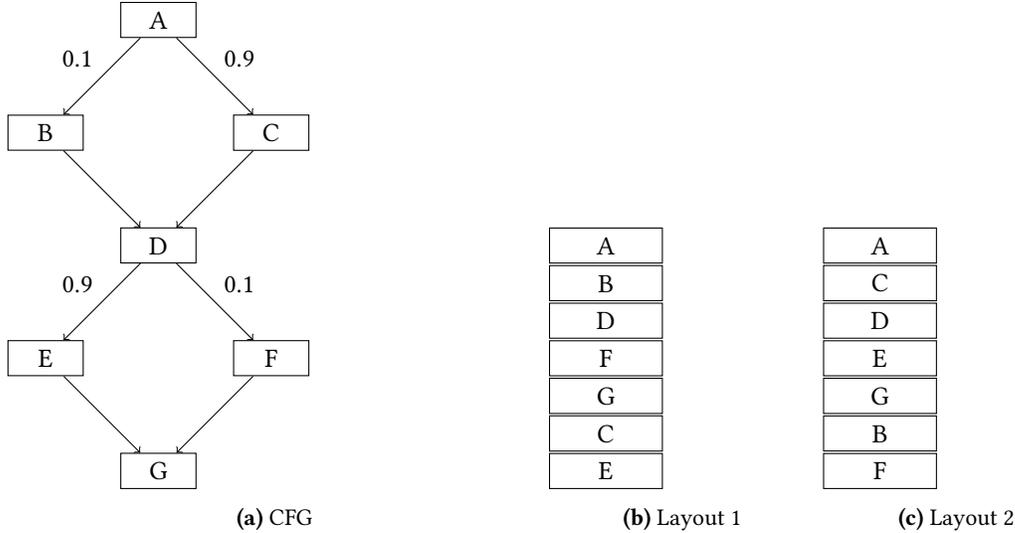
\begin{figure*}
    \centering
    \begin{subfigure}[b]{0.4\textwidth}

%    \subfloat[CFG \label{fig:layout_example2_a}]{
%    \centering
\tikzstyle{line} = [draw, ->]
\begin{tikzpicture}[auto,node distance = 15mm,box/.style = {draw,fill=white,align=center, minimum width=10mm}]
 \node[box] (b1)    {A};      
 \node[box, below of=b1, left of=b1] (b2)    {B};
 \node[box, below of=b1, right of=b1] (b3)    {C};  
 \node[box, below of=b2, right of=b2] (b4)    {D};     
  \node[box, below of=b4, left of=b4] (b5)    {E};
 \node[box, below of=b4, right of=b4] (b6)    {F};  
 \node[box, below of=b5, right of=b5] (b7)    {G};
 \path [line] (b1) -- node[swap] {0.1} (b2);
 \path [line] (b1) -- node {0.9} (b3);
 \path [line] (b2) -- (b4);
 \path [line] (b3) -- (b4);
 \path [line] (b4) -- node[swap] {0.9} (b5);
 \path [line] (b4) -- node {0.1} (b6);
 \path [line] (b5) -- (b7);
 \path [line] (b6) -- (b7);
 \end{tikzpicture}
 \caption{CFG}
 \label{fig:layout_example2_a}
\end{subfigure}
%}\qquad
%\subfloat[Layout 1 \label{fig:layout_example2_b}]{
\begin{subfigure}[b]{0.2\textwidth}
%\centering
\begin{tikzpicture}[auto,
  node distance = 5mm,
  %start chain = going below,
  box/.style = {draw,fill=white,
align=center, minimum width=15mm}]
 \node[box] (b1)    {A};      
 \node[box, below of=b1] (b2)    {B};
 \node[box, below of=b2] (b3)    {D};  
 \node[box, below of=b3] (b4)    {F};     
  \node[box, below of=b4] (b5)    {G};
 \node[box, below of=b5] (b6)    {C};  
 \node[box, below of=b6] (b7)    {E};
  \end{tikzpicture}
\caption{Layout 1}
\label{fig:layout_example2_b}
\end{subfigure}
%}\qquad
%\subfloat[Layout 2 \label{fig:layout_example2_c}]{
\begin{subfigure}[b]{0.2\textwidth}
%\centering
\begin{tikzpicture}[auto,
  node distance = 5mm,
  %start chain = going below,
  box/.style = {draw,fill=white,
align=center, minimum width=15mm}]
 \node[box] (b1)    {A};      
 \node[box, below of=b1] (b2)    {C};
 \node[box, below of=b2] (b3)    {D};  
 \node[box, below of=b3] (b4)    {E};     
  \node[box, below of=b4] (b5)    {G};
 \node[box, below of=b5] (b6)    {B};  
 \node[box, below of=b6] (b7)    {F};
  \end{tikzpicture}
  \caption{Layout 2}
\label{fig:layout_example2_c}
\end{subfigure}
%}\qquad    
    \caption{Code Layout}
    \label{fig:layout_example2}
\end{figure*}

 Code layout~\cite{ramirez:2001} is the process of ordering the blocks of the Control Flow Graph (CFG) linearly. This order dictates the placement of instructions within those blocks in memory. By inserting branch instructions at the end of the basic blocks, the compiler can layout the blocks in any order. Consider the Control Flow Graph (CFG) in Figure~\ref{fig:layout_example2_a}. Figures~\ref{fig:layout_example2_b} and \ref{fig:layout_example2_c} show two possible layouts of the CFG. In both theese layouts, a block is followed by one of its successor blocks that is not yet laid out. Consider block \textbf{A}, for example. It has two successors in the CFG. Only one of the successors -- \textbf{B} or \textbf{C} -- can be placed immediately following \textbf{A} and is known as the \textit{fall-through} block. In Figure~\ref{fig:layout_example2_b}, \textbf{B} is the fall-through block and in the layout in Figure~\ref{fig:layout_example2_c}, \textbf{C} is the fall-through block. The choice of which block to place as the fall-through block has performance implications. If control is often transferred to \textbf{C} from \textbf{A} often during program execution, then placing \textbf{C} next to \textbf{A} has the following advantages:
 
\begin{itemize}
    \item Since the branch at the end of \textbf{A} is mostly not-taken, the frontend of the processor's pipeline is less likely to be stalled if it is an out-of-order superscalar processor.
    \item As the cacheline containing the last instruction of \textbf{A} also contains instructions that are more likely to execute (from block \textbf{C}), instruction cache utilization is likely to be better.
\end{itemize}

% The following paragraph has an explicit line break.
In the LLVM compiler, the \texttt{MachineBlockPlacement} pass performs code layout optimization. This pass relies on the \emph{branch probability analysis} which provides, for each branch instruction, the probability of the branch being taken.\\ \texttt{MachineBlockPlacement} is just one of the many transformation passes that make use of branch probability analysis. Branch probability analysis is used in another analysis called \emph{block frequency analysis} that provides relative frequencies of basic blocks within a function. Block frequency analysis is used by optimizations such as inlining, spill-code placement in register allocation among others. While we use examples from machine block placement, this work on machine-learning based branch probability analysis has wider applicability within the compiler.

In PGO mode, branch probability analysis uses profile data in the form of \texttt{branch\_weights} metadata to derive the branch probabilities. However, there are many applications that are compiled without profile feedback and our work is targeted at this use case.

When a program is compiled without profile feedback, the compiler has very limited options in coming up with a reasonable branch  probability estimate.  
LLVM's branch probability analysis computes branch probabilities using these heuristics:
\begin{itemize}
    \item If the programmer has used \texttt{\_\_builtin\_expect} to specify the most likely outcome of the branch, the analysis uses that information.
    \item Loop back-edges are considered highly likely to be taken. Conversely, loop exit edges are considered highly unlikely to be taken.
    \item Certain comparisons are considered marginally less likely. These include comparing a pointer with \texttt{null} and equality comparison of floating point values among others.
    \item All other branches are considered to be unbiased (ie. both targets are considered equally likely).
\end{itemize}
% \begin{figure*}
%     \centering
% \begin{subfigure}[b]{0.3\textwidth}
% \begin{tikzpicture}[scale=0.55]
%  \pie [rotate = 180, explode={0, 0.2, 0}]
%     {42/NT,
%      10/Unbiased, 48/T}
% \end{tikzpicture}
%      \caption{Prediction: Unbiased}
% \end{subfigure}
% \begin{subfigure}[b]{0.3\textwidth}
% \begin{tikzpicture}[scale=0.55]
%  \pie [rotate = 90, explode={0, 0, 0.2}]
%     {20/Strongly NT,
%      40/Rest, 40/Strongly T}
% \end{tikzpicture}
%      \caption{Loop Back-edges}
% \end{subfigure}
% \begin{subfigure}[b]{0.3\textwidth}
% \begin{tikzpicture}[scale=0.55]
%  \pie [rotate = 270, explode={0.2, 0, 0}]
%     {44/Strongly NT,
%      31/Rest, 25/Strongly T}
% \end{tikzpicture}
%      \caption{Loop Exit-edges}
% \end{subfigure}
% \caption{Accuracy of Compiler Heuristics}
% \label{fig:static_accuracy}
% \end{figure*}

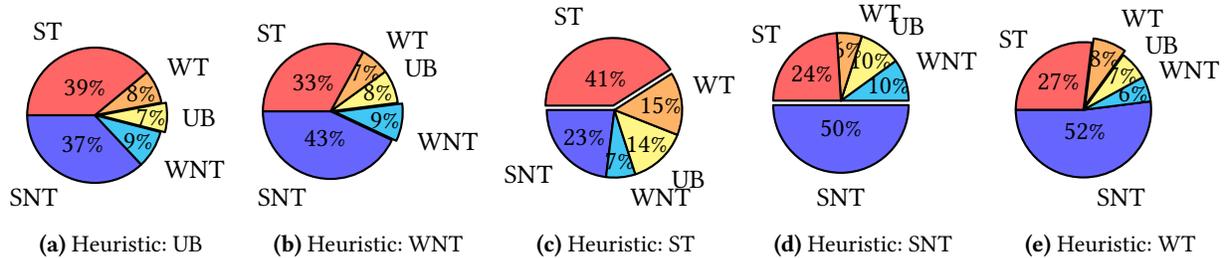
\begin{figure*}
    \centering
\begin{subfigure}[b]{0.18\textwidth}
\begin{tikzpicture}[scale=0.3]
 \pie [rotate = 180, explode={0, 0, 0.2, 0, 0}]
    {37/SNT, 9/WNT, 7/UB, 8/WT, 39/ST}
\end{tikzpicture}
     \caption{Heuristic: UB}
\end{subfigure}
\begin{subfigure}[b]{0.18\textwidth}
\begin{tikzpicture}[scale=0.3]
 \pie [rotate = 180, explode={0, 0.2, 0, 0, 0}]
    {43/SNT, 9/WNT, 8/UB, 7/WT, 33/ST}
\end{tikzpicture}
     \caption{Heuristic: WNT}
\end{subfigure}
\begin{subfigure}[b]{0.18\textwidth}
\begin{tikzpicture}[scale=0.3]
 \pie [rotate = 180, explode={0, 0, 0, 0, 0.2}]
    {23/SNT, 7/WNT, 14/UB, 15/WT, 41/ST}
\end{tikzpicture}
     \caption{Heuristic: ST}
\end{subfigure}
\begin{subfigure}[b]{0.18\textwidth}
\begin{tikzpicture}[scale=0.3]
 \pie [rotate = 180, explode={0.2, 0, 0, 0, 0}]
    {50/SNT, 10/WNT, 10/UB, 6/WT, 24/ST}
\end{tikzpicture}
     \caption{Heuristic: SNT}
\end{subfigure}
\begin{subfigure}[b]{0.18\textwidth}
\begin{tikzpicture}[scale=0.3]
 \pie [rotate = 180, explode={0, 0, 0, 0.2, 0}]
    {52/SNT, 6/WNT, 7/UB, 8/WT, 27/ST}
\end{tikzpicture}
     \caption{Heuristic: WT}
\end{subfigure}
\caption{Branches are categorized into 5 categories (\fb{Strongly} \fb{Not} \fb{Taken},  \fb{Weakly} \fb{Not} \fb{Taken}, \fb{Un}\fb{Biased}, \fb{Weakly} \fb{Taken}, and  \fb{Strongly} \fb{Taken}) based on their probability range. Caption under the chart shows how the compiler heuristic classifies the branches and chart shows the breakup of the branches with that classification. Charts are ordered in the descending order of number of branches with that heuristic categorization.}
\label{fig:static_accuracy}
\end{figure*}

% We collect the heuristic-based and profile-based probabilities of branches from an important production workload. We categorize the branches into three based on the three most common heuristic-based predictions: unbiased branches, loop back-edge branches predicted as strongly taken, loop exit-edge branches predicted as strongly not-taken. For branches falling under these three categories, we plot the cumulative distribution of their profile-based probabilities. This is shown in Figure~\ref{fig:static_histogram}. The X-axis in these graphs represent probabilities of branches and the Y-axis show the cumulative percentage of branches. The first plot on the left corresponds to branches that are predicted as unbiased by the compiler heuristics. As it can be seen, there are very few branches that are actually unbiased going by the profile data. In the other two categories, the static prediction fares better. Roughly half of the loop back-edge branches matches compiler heuristics with more than 90\% taken probability and roughly half of the loop-exit branches matches the heuristics with less than 10\% taken probability. Still there is a significant fraction of branches in both these categories whose behavior is quite the opposite of the prediction.
The GCC compiler also has very similar heuristics with some variations. We observe that these heuristics do not work well in practice by comparing the heuristic-based and profile-based probabilities of a large set of over 7 million branches collected from dataceneter workloads. We categorize the branches into five categories based on how they are predicted by LLVM heuristics: strongly not\-taken, weakly not\-taken, unbiased, weakly taken, and strongly taken. Branches belonging to each of these five categories are further categorized based on their profile\-based probabilities. This is shown in the pie charts in Figure~\ref{fig:static_accuracy}.  The first chart on the left corresponds to branches that are predicted as unbiased by the compiler heuristics. Only 7\% of these branches categorized as unbiased based on static heuristics are actually unbiased. Even when the categorization is better such as in the case of the strongly taken and not\-taken branches, there is still a significant fraction of branches with diametrically opposite behavior.

This illustrates the main limitation of compiler heuristics: heuristics need to be kept simple in order to keep the code readable and maintainable. The very simplicity results in branches the default choice -- equal probabilities -- for a significant fraction of branches. 

This motivates us to explore the use of machine learning to address this problem. 

% \begin{figure*}
%     \centering
%     \includegraphics[scale=0.8]{static_histogram.pdf}
%     \caption{Accuracy of Static Prediction}
%     \label{fig:static_histogram}
% \end{figure*}

\section{Machine Learning Model for Branch Probability Estimation}
\label{sec:training}

\subsection{Machine Learning Overview}
We now present a brief overview of the machine learning terminology~\cite{murphy:2012,MLGlossary} that are used in this work. A machine learning system builds a predictive \textbf{model} from input data. The input data consists of a set of \textbf{examples}. The flavor of machine learning we use in this work is \textbf{supervised machine learning}. In a supervised machine learning system, an example can be a \textbf{labeled example} or an \textbf{unlabeled example}. Each labeled example consists of a set of \textbf{features} and a \textbf{label}. An unlabeled example consists of a set of features without any label. In this problem of estimating branch  probabilities, each example describes an LLVM IR-level branch instruction. The features of a branch instruction are described in detail later in this section. The label for a branch instruction is the probability of the branch being taken and is obtained from profiling data. \textbf{Training} is the process of building an ML model from a set labeled of examples. After a model is trained, it is used to predict the labels of unlabeled examples. This process is called \textbf{inference}. 

Supervised machine learning problems can be further classified into \textbf{regression} or \textbf{classification} problems. In a regression problem, the ML model predicts a numeric quantity. Branch probability prediction is an example of a regression problem as the output is a numeric value between 0 to 1.

\begin{figure*}
    \includegraphics[scale=0.6]{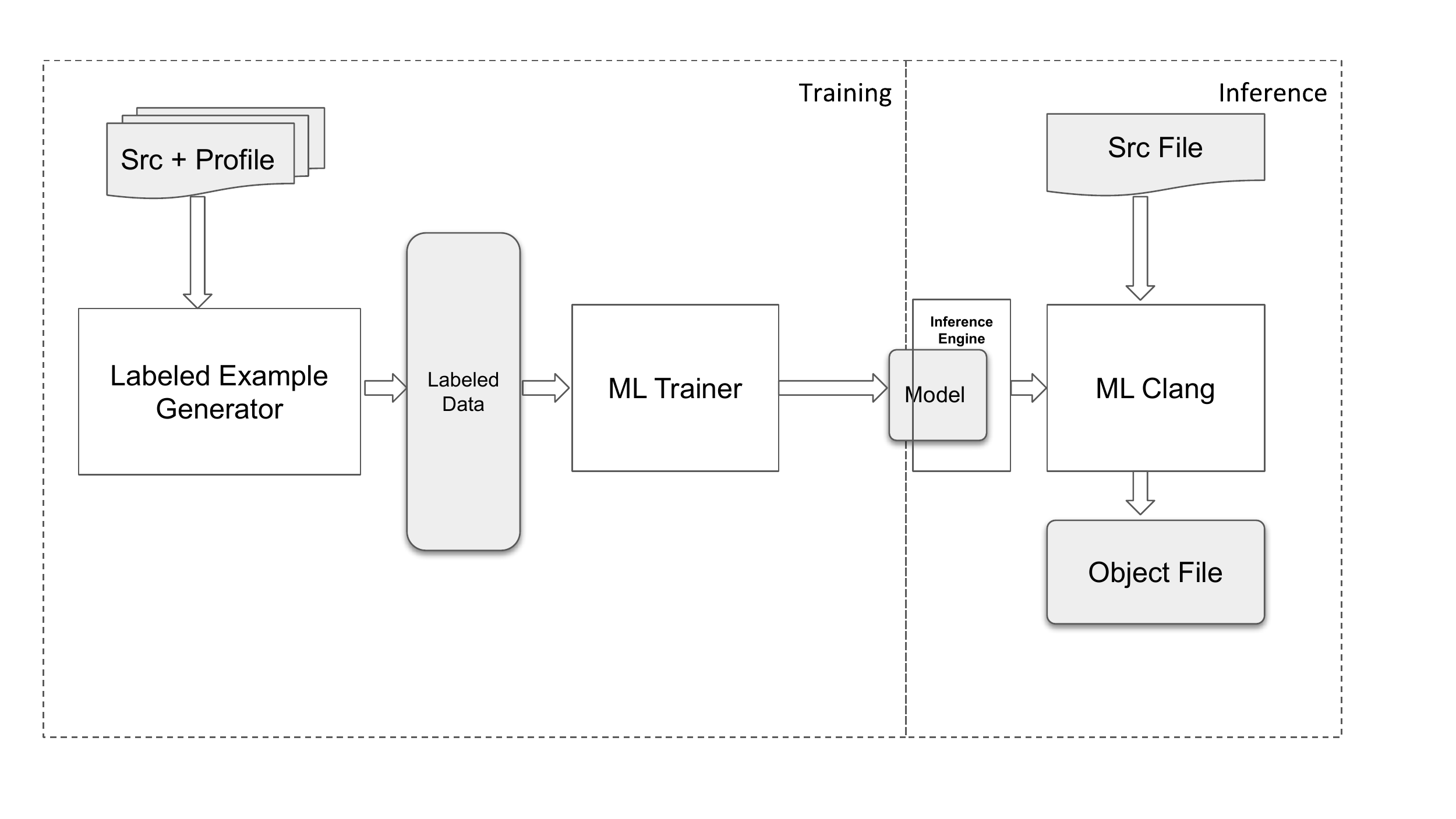}
    \caption{Machine Learning Overview}
    \label{fig:ml_diagram}
\end{figure*}

Figure~\ref{fig:ml_diagram} depicts how our machine learning based system works. The system consists of a labeled example generator that produces the training data, an ML trainer that produces a trained model and an inference engine that is integrated with the Clang compiler.
On the training side, the labeled example generator we use is a Clang binary with modifications. This Clang builds multiple build targets, where each target consists of multiple source files. This compilation also takes in a sample profile file which was generated using Google Wide Profiling infrastructure~\cite{ren:2010}. This Clang compiler extracts features for each conditional branch, matches the sample profile data for those branches, and produces a set of labeled examples. Labeled examples from multiple compilations are aggregated and fed to a ML training process. This produces a trained model. On the inference side, the Clang binary is enhanced to integrate with an inference engine that takes the trained model. The inference engine could either be in-process or could be a separate service. 

\subsection{Why Machine Learning?}
It may not be obvious that machine learning is the right tool to address this problem. The math behind supervised machine learning is based on the assumption that the data used to train the model to be independent and identically distributed. This is clearly not the case here as the branches within a program are correlated. While supervised machine learning has been used in other instances where the data is not \emph{iid}, the absence of \emph{iid} property could have an impact on the efficacy of the approach. In spite of this, ML is a promising approach for these reasons:

\begin{itemize}
\item The combination of Google Wide Profiling~\cite{ren:2010} and SamplePGO~\cite{chen:2016} provides us with  probabilities for a large number of branches.
\item One can use ML to compute a complex function of its inputs without having that complexity come in the way of code maintainability. 
\item ML models can take a larger set of features as input.
\item Codebase-specific and workload-specific training of ML models is possible.
\end{itemize}

\subsection{Training}
In this section, we describe the components of the box labeled \texttt{Training} in Figure~\ref{fig:ml_diagram}. The training component consists of a labeled examples generator that produces labeled examples when provided with source files and profile data, and an ML trainer that takes these examples and produces a trained model. Generating a large set of labeled examples is the first step of the training process. In our implementation, we use a modified Clang binary as the labeled examples generator, but this could also be implemented as a standalone tool using LLVM libraries. We compile a set of build targets representing some of the most important applications that run in Google Datacenters using using the labeled example generator. Each target has an associated SamplePGO profile file that is generated by the GWP pipeline. A single source file may be compiled multiple times if it belongs to multiple build targets but each (source file, profile) pair is compiled once.

The modifications to Clang to transform it into a labeled example generator happen in the \texttt{SampleProfileLoader} pass in the LLVM pass pipeline. This pass first matches profile data with LLVM IR  using debug information and then computes profile weights of basic blocks contained in the profile. After basic block annotation, profile weights are propagated to other blocks using control equivalence information. Finally, profile weights for control flow edges are computed using network flow constraints.  After it annotates the branches with edge weights, we extract the set of features for each branch, its label and its sample count. This information is dumped into a file in CSV format. We also collect the probabilities of the branch as estimated by the compiler heuristics. This is only for the purpose of comparing the heuristic driven and ML generated approaches, and is not used to train the model.

\subsubsection{Features}
\label{subsec:features}

Compiler heuristics for branch probability estimation use simple combinations of branch attributes. These attributes are what are referred to as \emph{features} in ML terminology.
 For example, whether the branch is a back-edge of a loop is a feature and the compiler heuristics  consider a branch to be highly taken if the value of that feature is True. A machine learning model, on the other hand, can be fed more features and the way they are combined is opaque to the model developer. Choosing the right set of features is a critical step in the success of an ML model. On one extreme, we could use the very limited set of features used in the current compiler heuristics. While the ML model may do a better job in combining them we still miss signals that might be useful in predicting the branch. On the other extreme, the entire program in LLVM IR could be used as the feature set. While this could in theory not miss any signals, this requires ML advances to understand the structure of the program to derive useful signals from it. The approach we take in this work is to use the features used in current compiler heuristics as a starting point and generalize and expand them based on intuition.   

Some of the features (loop depth, for example) are numeric features while others (the opcodes, for example) are categorical features. Numeric values are \emph{standardized} by subtracting the mean of the distribution from each data point and dividing by the standard deviation ensuring they have a mean of 0 and standard deviation of 1. Standardization ensures that two features whose typical values are vastly different are treated on an equal footing.

Categorical values are handled in two ways. If a feature has only a few possible values, then they are encoded using \emph{one-hot encoding}. One-hot encoding represents a value with N possible categories as a sparse vector of N elements. Each index in the vector corresponds to one category and has a 1 if the value to be encoded belongs to that category and 0 if the value does not belong to the category. For example, if a categorical feature has two possible values \texttt{cat} and \texttt{dog}, one-hot encoding could represent \texttt{cat} as [1, 0] and \texttt{dog} as [0, 1].

If a categorical feature has a large number of categories, using one-hot encoding increases the dimensionality of the input. In such cases, a technique called \emph{embedding} is used. Embedding translates a high-dimensional sparse vector to a low-dimensional dense vector. Later in this section, we describe a categorical feature (the names of the functions called in control dependent blocks), with a large number of categories. To encode this, we first replace the strings by integers that are indices to a table of such strings. These indices are then embedded into low-dimensional dense vectors.

\paragraph{Dataflow Features:}

\begin{figure}
    \centering
\begin{tikzpicture}[sibling distance=10em,
  every node/.style = {shape=rectangle, rounded corners,
    draw, align=center},
%    top color=white, bottom color=blue!20},
    level 1/.style={sibling distance=3cm,level distance=1.4cm},
    level 2/.style={sibling distance=1cm},
    level 3/.style={sibling distance=1cm}]
    
  \node {icmp eq}[grow'=up]
    child { node {*} 
      child {node {VAR} }
      child {node {3} }}
    child { node {+} 
      child {node {VAR}} 
      child {node {2} }
      };
     
\end{tikzpicture}
\caption{Expression tree}
\label{fig:expr_tree}

\end{figure}
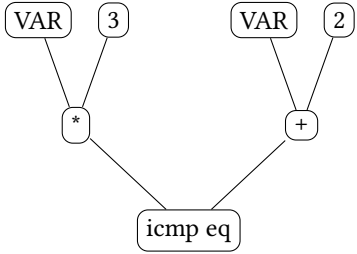

The instructions that produce the branch condition play a role in the outcome of the branch. We capture this by a set of features that are derived from the backward dataflow of the branch condition. This is an extension and generalization of heuristics such as \textit{equality comparison of null and a pointer is less likely to be true}.  Consider the following sequence of LLVM IR-level instructions:

\begin{lstlisting}
%10 = %9 + 2
%11 = %8 * 3
%12 = icmp eq, %11, %10
br %12, label %t, label %f

\end{lstlisting}

Starting from the branch condition \%12, we identify the backward dataflow tree of height 2 feeding into \%12 and encode the nodes of that tree with constant or operators. For the above example, the dataflow tree looks like the one in Figure~\ref{fig:expr_tree}. Nodes in the tree are operators, constants, and a marker node \texttt{VAR} representing variables. The seven nodes of this tree are encoded and added to the feature set. Operators are represented by their opcodes. Most integer constant operands are encoded using the actual constants. If the constant exceeds a threshold, it is marked simply as a constant of unknown value. Similarly, a special value is used to encode all variables. We do not distinguish between the variables as the actual variable name plays no part in determining the branch outcome.

\paragraph{Control Flow Features:}

A block \texttt{B} is said to be \emph{control dependent} on a control flow edge \texttt{E} if \texttt{B} is executed \emph{only} if the edge \texttt{E} is taken. Consider the following LLVM IR fragment:

\begin{lstlisting}
%12 = icmp eq, %11, 0
br %12, label %t, label %f
%t:
call void @log_error
\end{lstlisting}

The block starting from the label \%t is control dependent on the taken edge of the branch and it has a call to \texttt{log\_error}. If \texttt{log\_error} is called very infrequently, we can infer that the branch is highly unlikely to be taken. To capture this, we record the most frequently occuring function call in the control dependent blocks of the taken and not-taken edges of the branch. We apply embedding to reduce the dimensionality of this feature. In addition to the function names, we also record certain  properties of these functions. These properties include \texttt{inline} or \texttt{noinline} keywords and attributes such as \texttt{\_always\_inline\_} and \texttt{cold}.

 We also encode the ''shape'' of the control flow graph starting from the source block(the block containing the branch). We only consider simple shapes like a triangle (the block has two successors \texttt{A} and \texttt{B} with \texttt{A} having \texttt{B} as its sole successor) and diamond (the block has two successors \texttt{A} and \texttt{B}, both having \texttt{C} as their sole successor)

\paragraph{Loop Features:}
A common heuristic used in compilers is loop back-edge branches are strongly taken and loop exit branches are strongly not-taken. Looking at the profile data of a large set of branches, we noticed that while this is a good heuristic in general, there are non-trivial exceptions. The intention behind adding properties of the enclosing loop to the feature set is to let the ML system differentiate low trip-count loops from high trip-count loops.

We use the following numeric features related to the loop enclosing the branch:
\begin{itemize}
    \item Loop depth
    \item Number of basic blocks in the loop
    \item Number of exit blocks of the loop. These are blocks outside of the loop that are successors of any branch inside the loop.
    \item Number of exit edges in the loop
\end{itemize}

In addition, a categorical feature relates the branch to its enclosing loop. For each of the two outgoing edges from the branch, we assign one of the following four categories:

\begin{itemize}
    \item The edge is a loop exit edge
    \item The edge is a loop back-edge
    \item The edge destination is the header of an inner loop
    \item The edge destination is within the same loop
\end{itemize}

The last category also applies to branches which do not have an enclosing loop.
\paragraph{Function Features:}
Properties of enclosing function including the number of instructions, number of basic blocks and the number of edges in the control-flow graph of the function are part of the feature set.

\paragraph{File Name:}
Consider a conditional branch in a header file of a library. As this header file gets included in multiple source files, the branches in the header file gets replicated. Do these branches behave similarly even though the surrounding contexts are different? In order to explore that, we extract the file name from the debug location of the branch instruction and add it to the feature set. This does present the risk of over-fitting to a specific codebase, but one of the benefits of an ML based approach is that a model can be specifically trained for a specific large codebase. Furthermore, we only use the file name and not specific line numbers as the churn in codebase causes line numbers to frequently change where as the file names of widely used libraries are fairly stable.

\subsubsection{Labels}
In a supervised learning system, training the model requires the input to be labeled examples. The label of a branch is the probability that the branch is taken. SamplePGO annotates branch instructions in LLVM IR with \texttt{branch\_weights} metadata which contains the taken(\texttt{t}) and non-taken(\texttt{nt}) weights of the branch. From this metadata, we obtain the taken probability as \texttt{t/(t+nt)}. 

\subsection{ML Trainer}

 We use a neural network to compute branch probabilities from input features. A neural network consists of a sequence of \emph{layers}. There is an input layer, an output layer, and one or more \emph{hidden layer}s that are also known as \emph{dense layer}s. Each layer of a neural network consists of a set of nodes representing artificial neurons. 

Our machine learning model is implemented as a neural network in TensorFlow~\cite{abadi:2015} using its Keras APIs. We use 5 hidden layers in our model. 
Neural networks with more than one hidden layers are usually referred to as \emph{deep neural networks}. In Section~\ref{sec:results} we discuss the effects of the number of hidden layers in our model. 

A dense layer in the neural network first computes a linear function of its inputs and then applies a non-linear \emph{activation function}. A common activation function that is used is the \emph{rectifier function} usually referred to as \texttt{ReLU}. A rectifier function f(x) is defined as f(x) = max(0, x). We use \texttt{ReLU} for all our hidden layers.  The final layer uses \texttt{sigmoid} activation to generate a floating point number in the [0, 1] range.

The edges connecting the neurons have \emph{weights} and \emph{biases} associated with them. These form the \emph{learnable parameters} of the neural network. These learnable parameters, along with the activation functions, determine how the inputs get transformed into the output value. The goal of training is to assign values for these parameters so as to minimize the value of a \emph{loss function}. A discussion on loss functions is presented later in this section.

At a high level, the trainer computes the loss for the training input and updates the learnable parameters using \emph{gradient descent} methods. There are several \emph{optimizers} used to optimize gradient descent. We use the \emph{adaptive gradient} or \textbf{Adagrad} optimizer~\cite{duchi:2011}. The training algorithm computes the loss for a \emph{batch} of examples at a time. We use a batch size of 200 in our training. The processing of the entire training set once by the training algorithm constitutes one \emph{epoch}. Our model is trained for 100 epochs.

\begin{table}
\centering
\caption{Size of Training Data}
\label{table:training_stats}
 \begin{tabular}{ll} 
 \toprule
 Attribute & Quantity \\
 \midrule
 \# Build Targets & \textgreater 750 \\
 \# Source Files & \textgreater 37000\\
 \# Branches &  \textgreater 40M \\
 \# Unique Examples &  \textgreater 7M \\
 \bottomrule
 \end{tabular}

\end{table}

Table~\ref{table:training_stats} shows some metrics illustrating the size of our training data. We build more than 750 build targets from a shared codebase compiling more than 37K C/C++ source files. This results in labeled data for 40M static branch instructions and more than 7M unique examples. There are two sources of repetition: branches in a single source file being part of multiple targets and different branches with the same exact set of (features, label) pairs. As a comparison, the training data used by Calder et al.~\cite{calder:1996} contains a total of around 55K branches, but they do not specify the number of unique branches. 

\subsubsection{Loss Functions}
The process of training a neural network model involves finding the weights of the neural network so as to minimize a \emph{loss function}. We discuss three different relevant loss functions that we considered for this problem:

\paragraph{Mean Absolute Error:} Mean Absolute Error(MAE) computes the mean of absolute difference between the predicted and actual values. 

\paragraph{Mean Squared Error:} The difference between the predicted and the actual values are squared and their mean is taken as Mean Squared Error(MSE).

\paragraph{Cross Entropy Loss:} In information theory, \emph{entropy} refers to the optimal number of bits needed to encode values from a set. This depends on the probability distribution of those values: a value that appears more frequently is encoded using fewer bits than a value that occurs less frequently, thereby  reducing the weighted average. The ML model does not know the true probability distribution of values $\mathbb{P}$ and instead approximates it with a different probability distribution $\mathbb{Q}$. Let us use $\mathbb{Q}$ to identify the bits needed to encode the values from the set but compute the weighted average using $\mathbb{P}$. This weighted average, computed using $\mathbb{P}$ for the weights and $\mathbb{Q}$ for the number of bits  is known as \emph{cross entropy}.

Based on our experiments we find cross entropy loss to generate models that improve the application performance better than the other two loss functions we considered.

\subsection{Inference}
The inference engine depicted in Figure~\ref{fig:ml_diagram} is implemented in-process in Clang.  Inference is done inside a new LLVM pass named \texttt{MLBranchProbAnnotate}. This pass can be inserted in the LLVM pass pipeline anywhere before lowering the IR to LLVM's \texttt{MachineIR}. This pass loads the given ML model at start-up. For each conditional branch instruction in the IR, it gathers the set of features for that branch. The features are converted to Tensors. Numeric features are standardized using the mean and standard deviation obtained from the training data. The standardized input is fed to the \texttt{SavedModel} APIs of TensorFlow to obtain the label, which is the probability of the taken side of the branch. This probability is converted to branch weights which is then annotated to the branch instruction using LLVM's \texttt{MDProf} metadata.

\section{Evaluation}
\label{sec:results}
%\begin{figure}
%    \centering
%    \includegraphics[scale=0.38]{chart.pdf}
%    \caption{Prediction Accuracy of ML}
%    \label{fig:prediction-accuracy}
%\end{figure}

\pgfplotstableread[row sep=\\,col sep=&]{
    Metric & Heuristics & ML \\
    RMSE     & 0.44  & 0.32   \\
    MAE     & 0.38 & 0.19   \\
    Cross Entropy    & 0.79 & 0.50 \\
    Closeness & 0.32 & 0.68 \\
    }\searchstats
 
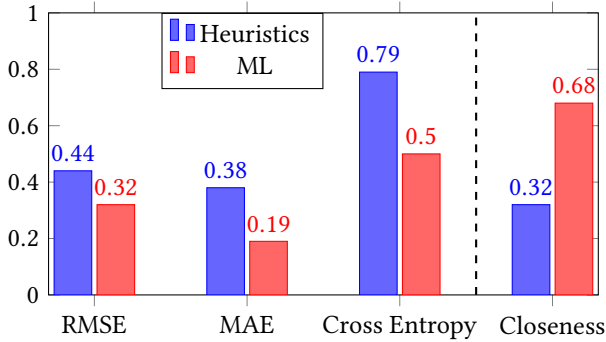
\begin{figure}
    \centering

\begin{tikzpicture}
    \begin{axis}[
            ybar,
            bar width=.5cm,
            width=0.5\textwidth,
            height=.3\textwidth,
            legend style={at={(0.5,1)},
                legend columns=1},
            symbolic x coords={RMSE, MAE, Cross Entropy, Closeness},
            xtick=data,
            nodes near coords,
            ymin=0,ymax=1
        ]
        \addplot[blue,fill=blue!60!white] table[x=Metric,y=Heuristics]{\searchstats};
        \path
    (axis cs:Cross Entropy, \pgfkeysvalueof{/pgfplots/ymin})
    -- coordinate (tmpmin)
    (axis cs:Closeness, \pgfkeysvalueof{/pgfplots/ymin})
    (axis cs:Cross Entropy, \pgfkeysvalueof{/pgfplots/ymax})
    -- coordinate (tmpmax)
    (axis cs:Closeness, \pgfkeysvalueof{/pgfplots/ymax})
  ;
    \draw[thick, dashed] (tmpmin) -- (tmpmax);
        \addplot[red,fill=red!60!white] table[x=Metric,y=ML]{\searchstats};
        \legend{Heuristics, ML}
    \end{axis}
\end{tikzpicture}
    \caption{Aggregate Metrics on 10\% of all branches. Higher values are better for \texttt{Closeness} and lower values are better for all other metrics.}
    \label{fig:agg_metrics_ten_percent}
\end{figure}    

% Need to collect Cross Entropy for Heuristics    

\pgfplotstableread[row sep=\\,col sep=&]{
    Metric & Heuristics & ML \\
    RMSE     & 0.46  & 0.36   \\
    MAE     & 0.41 & 0.26   \\
    Cross Entropy    & 0.77 & 0.52 \\
    Closeness & 0.33 & 0.67 \\
    }\searchstats
 
\begin{figure}
    \centering

\begin{tikzpicture}
    \begin{axis}[
            ybar,
            clip=false,
            bar width=.5cm,
            width=0.5\textwidth,
            height=.3\textwidth,
            legend style={at={(0.5,1)},
                legend columns=1},
            symbolic x coords={RMSE, MAE, Cross Entropy, Closeness},
            xtick=data,
            nodes near coords,
            ymin=0,ymax=1
        ]
        \addplot[blue,fill=blue!60!white]  table[x=Metric,y=Heuristics]{\searchstats};
        %\node[yshift=-1.0cm] at (axis cs:MAE,0) {(smaller is better)};
          \path
    (axis cs:Cross Entropy, \pgfkeysvalueof{/pgfplots/ymin})
    -- coordinate (tmpmin)
    (axis cs:Closeness, \pgfkeysvalueof{/pgfplots/ymin})
    (axis cs:Cross Entropy, \pgfkeysvalueof{/pgfplots/ymax})
    -- coordinate (tmpmax)
    (axis cs:Closeness, \pgfkeysvalueof{/pgfplots/ymax})
  ;
    \draw[thick, dashed] (tmpmin) -- (tmpmax);
        \addplot[blue,fill=red!60!white]  table[x=Metric,y=ML]{\searchstats};
        \legend{Heuristics, ML}
    \end{axis}
\end{tikzpicture}
    \caption{Aggregate Metrics on branches from \texttt{search}. Higher values are better for \texttt{Closeness} and lower values are better for all other metrics.}
    \label{fig:agg_metrics_search}
\end{figure}
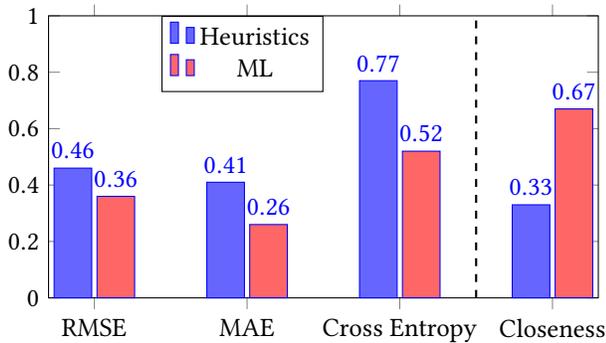

    % \begin{figure}
    %     \centering
    %     \includegraphics[scale=0.3]{error_cdf_ten_percent.pdf}
    %         \caption{Prediction Error}

    %     \label{fig:error_cdf_ten_percent}
    % \end{figure}
    %     \begin{figure}
    %     \centering
    %     \includegraphics[scale=0.3]{error_cdf_search.pdf}            
    %     \caption{Prediction error: Search}

    %     \label{fig:error_cdf_search}
    % \end{figure}

 \pgfplotstableread[row sep=\\,col sep=&]{
Error & 0HL & 1HL & 2HL & 3HL & 4HL & 5HL & Heuristics \\
0 & 1.753433489 & 2.806879137 & 3.562854695 & 4.001425142 & 4.206289287 & 4.316143973 & 3.717810598 \\
1 & 3.494990796 & 5.968205763 & 7.656461633 & 8.369880871 & 8.794878538 & 9.097155654 & 6.215060696 \\
2 & 5.452722898 & 9.244901018 & 11.53007643 & 12.66481738 & 13.38035736 & 13.72773569 & 9.371580296 \\
3 & 7.418089688 & 12.13632726 & 14.9649794 & 16.48555489 & 17.13252406 & 17.85386075 & 12.0003167 \\
4 & 9.455703254 & 14.94391332 & 18.22711219 & 19.83647628 & 20.66102265 & 21.43608493 & 13.29241706 \\
5 & 11.26470031 & 17.52514499 & 21.18399882 & 22.70740197 & 23.66513125 & 24.48501198 & 14.96992781 \\
6 & 12.88226462 & 19.79123368 & 23.53604845 & 25.1499368 & 26.15616049 & 27.14810448 & 15.673733 \\
7 & 14.44299291 & 21.89388348 & 25.70062407 & 27.2634731 & 28.45208102 & 29.53535851 & 16.32423674 \\
8 & 15.95550956 & 23.91693459 & 27.71844399 & 29.29458305 & 30.57537276 & 31.81360615 & 16.9152182 \\
9 & 17.3583129 & 25.84681421 & 29.65199958 & 31.33276215 & 32.73782762 & 34.08054314 & 17.47410568 \\
10 & 18.70555268 & 27.68960184 & 31.56321664 & 33.43088769 & 34.88741662 & 36.286544 & 18.00160611 \\
11 & 20.1387534 & 29.51584763 & 33.38479679 & 35.36557434 & 36.76597417 & 38.18206753 & 18.58537703 \\
12 & 21.41643268 & 31.15730569 & 35.00052312 & 37.14530509 & 38.5500878 & 40.02075504 & 19.31392426 \\
13 & 22.73624413 & 32.76935606 & 36.6384466 & 38.88474165 & 40.35060187 & 41.8598667 & 19.9116921 \\
14 & 24.00586457 & 34.47090194 & 38.26590772 & 40.59674988 & 42.11407356 & 43.66872238 & 20.46831746 \\
15 & 25.28043342 & 36.00278808 & 39.90340705 & 42.33986241 & 43.83385786 & 45.38681009 & 20.98295202 \\
16 & 26.60957616 & 37.55644721 & 41.57497971 & 43.99913474 & 45.50048212 & 47.0252991 & 21.54848451 \\
17 & 28.03867677 & 39.17118386 & 43.2370797 & 45.64539982 & 47.10390811 & 48.71765508 & 22.05407056 \\
18 & 29.43525925 & 40.69147658 & 44.88079989 & 47.23963591 & 48.72656221 & 50.33649183 & 22.59669899 \\
19 & 30.8380626 & 42.22746282 & 46.48422588 & 48.83486168 & 50.33635045 & 51.93030378 & 23.16887648 \\
20 & 32.30986091 & 43.80798475 & 48.04170237 & 50.34568174 & 51.92097249 & 53.511674 & 23.69142851 \\
21 & 33.74956525 & 45.40716926 & 49.606248 & 51.87855755 & 53.43872031 & 55.08696476 & 24.1907937 \\
22 & 35.16481031 & 46.97892543 & 51.11791635 & 53.41609901 & 54.93950216 & 56.56173211 & 24.79195473 \\
23 & 36.65965406 & 48.56467854 & 52.60003563 & 54.89100775 & 56.45244296 & 58.02504743 & 25.39467099 \\
24 & 38.2049716 & 50.07069156 & 54.09403109 & 56.38387214 & 57.89624741 & 59.42502312 & 25.95454815 \\
25 & 39.81080111 & 51.54121742 & 55.5540946 & 57.84124937 & 59.32888259 & 60.83574392 & 26.63445959 \\
26 & 41.31752104 & 53.04793736 & 57.10322947 & 59.32336865 & 60.78640121 & 62.27191368 & 27.30701911 \\
27 & 42.87895625 & 54.59891022 & 58.56442405 & 60.73946201 & 62.20772574 & 63.65350955 & 27.97816479 \\
28 & 44.58870236 & 56.12754454 & 59.97952772 & 62.17492485 & 63.63922986 & 65.03821586 & 28.6993601 \\
29 & 46.23708819 & 57.65476504 & 61.48568213 & 63.57659713 & 64.99919412 & 66.343889 & 29.73555701 \\
30 & 47.98161454 & 59.18354074 & 62.88594058 & 64.93373373 & 66.32550919 & 67.59413995 & 30.8380626 \\
31 & 49.78849085 & 60.68601919 & 64.31362735 & 66.31462269 & 67.65253118 & 68.87563658 & 31.92798509 \\
32 & 51.51675814 & 62.14806206 & 65.71289612 & 67.67529386 & 69.017868 & 70.16476789 & 33.22658908 \\
33 & 53.26835365 & 63.60444961 & 67.16221451 & 69.00415384 & 70.2811262 & 71.36949348 & 34.68919748 \\
34 & 55.03267364 & 65.08854825 & 68.5571004 & 70.29088164 & 71.5125732 & 72.54071127 & 36.47288696 \\
35 & 56.83544984 & 66.52952504 & 69.88087058 & 71.53745663 & 72.71701602 & 73.6973666 & 38.6916123 \\
36 & 58.76858128 & 68.00641314 & 71.20775119 & 72.78657652 & 73.89092009 & 74.81853476 & 41.81745177 \\
37 & 60.791491 & 69.5112951 & 72.51568646 & 74.008975 & 75.05690671 & 75.9315027 & 46.74507775 \\
38 & 62.64078224 & 70.99129363 & 73.76636156 & 75.17778928 & 76.21200682 & 77.03457383 & 47.4340377 \\
39 & 64.41443352 & 72.45658831 & 75.01944018 & 76.34306898 & 77.31309858 & 78.10823726 & 48.10179019 \\
40 & 66.26301785 & 73.85119144 & 76.26714624 & 77.52630433 & 78.38619648 & 79.15489652 & 48.77010821 \\
41 & 68.11216771 & 75.22373879 & 77.51683166 & 78.65100707 & 79.47371546 & 80.18458981 & 49.9656439 \\
42 & 70.01447763 & 76.60561743 & 78.76368942 & 79.80893485 & 80.5683036 & 81.20141722 & 51.56921128 \\
43 & 71.95354716 & 78.01351057 & 79.98721897 & 80.92218556 & 81.57169962 & 82.15928788 & 53.10590444 \\
44 & 73.78389307 & 79.40175145 & 81.15433664 & 81.94042681 & 82.51401814 & 83.04165995 & 55.0263114 \\
45 & 75.60391801 & 80.6997899 & 82.24298669 & 82.90126651 & 83.43272567 & 83.88854486 & 57.55353472 \\
46 & 77.30758464 & 81.96842067 & 83.27861806 & 83.83637448 & 84.33446723 & 84.72581571 & 60.4404367 \\
47 & 78.91709011 & 83.15801826 & 84.27296557 & 84.74843701 & 85.19690428 & 85.54032388 & 64.02548855 \\
48 & 80.6160911 & 84.2718345 & 85.22419122 & 85.64183696 & 86.03530619 & 86.33645225 & 69.06523135 \\
49 & 82.17908152 & 85.3491739 & 86.14615056 & 86.52929882 & 86.81531688 & 87.1050109 & 80.48601862 \\
50 & 83.66049388 & 86.38169484 & 87.0405402 & 87.33687922 & 87.58528937 & 87.8321443 & 82.35793117 \\
51 & 85.05947988 & 87.35921776 & 87.89053553 & 88.14431824 & 88.33009566 & 88.54584631 & 82.59078917 \\
52 & 86.38240176 & 88.33589237 & 88.72158553 & 88.90806987 & 89.05143235 & 89.21981965 & 82.84598571 \\
53 & 87.70998928 & 89.24145127 & 89.51036197 & 89.62997209 & 89.73360592 & 89.88474448 & 83.05396028 \\
54 & 89.00251379 & 90.09300182 & 90.28528287 & 90.33646356 & 90.40192394 & 90.5161615 & 83.57396741 \\
55 & 90.10261587 & 90.89266476 & 91.01920266 & 91.02853394 & 91.05341737 & 91.13683341 & 84.09128825 \\
56 & 91.15775812 & 91.61230486 & 91.71028336 & 91.68879313 & 91.6822895 & 91.73898413 & 84.68057311 \\
57 & 92.11265973 & 92.28387469 & 92.34721433 & 92.29023693 & 92.24909444 & 92.29433704 & 85.37221935 \\
58 & 92.99149722 & 92.90638458 & 92.90836394 & 92.85166931 & 92.77447413 & 92.80854746 & 86.28611985 \\
59 & 93.82141615 & 93.50627317 & 93.43925757 & 93.37422133 & 93.30282286 & 93.31144723 & 87.23437646 \\
60 & 94.53271464 & 94.06247437 & 93.94145042 & 93.85464118 & 93.78479792 & 93.77560802 & 88.72328212 \\
61 & 95.18463222 & 94.56961564 & 94.41607356 & 94.31229835 & 94.24174817 & 94.2121991 & 91.2257634 \\
62 & 95.70873946 & 95.05851847 & 94.85987519 & 94.7583621 & 94.67692543 & 94.62136186 & 94.89267607 \\
63 & 96.16272066 & 95.49256466 & 95.28770052 & 95.17063529 & 95.09131936 & 95.01525524 & 94.94314985 \\
64 & 96.59153568 & 95.87359218 & 95.69262178 & 95.570184 & 95.48351614 & 95.38525487 & 94.9854234 \\
65 & 96.96931138 & 96.24401596 & 96.0819909 & 95.94103193 & 95.84630523 & 95.75553727 & 95.03264536 \\
66 & 97.30368246 & 96.58489067 & 96.42272423 & 96.28685505 & 96.20555975 & 96.09966379 & 95.08099839 \\
67 & 97.6384777 & 96.90173025 & 96.74140179 & 96.60468431 & 96.5201372 & 96.40971698 & 95.11719247 \\
68 & 97.9191232 & 97.20923854 & 97.02727846 & 96.89466109 & 96.81534516 & 96.71185271 & 95.15395208 \\
69 & 98.16951271 & 97.49553936 & 97.30113757 & 97.18336543 & 97.0889215 & 96.97991511 & 95.19580149 \\
70 & 98.38370814 & 97.74875654 & 97.56835167 & 97.45991082 & 97.34977336 & 97.23779793 & 95.2409027 \\
71 & 98.59719666 & 97.97779719 & 97.80375457 & 97.68569966 & 97.58687286 & 97.48380456 & 95.28515562 \\
72 & 98.76515981 & 98.1863373 & 98.02728129 & 97.9199715 & 97.80177521 & 97.719773 & 95.32347045 \\
73 & 98.91446038 & 98.36829738 & 98.21404839 & 98.12158383 & 98.02162596 & 97.92152671 & 95.36235081 \\
74 & 99.04877435 & 98.55068161 & 98.39812922 & 98.30368529 & 98.20627232 & 98.10447647 & 95.40321053 \\
75 & 99.16838447 & 98.72316902 & 98.55124714 & 98.47235536 & 98.37720452 & 98.28671932 & 95.44901866 \\
76 & 99.27908746 & 98.88194226 & 98.69913389 & 98.62914924 & 98.54912639 & 98.45906534 & 95.49525094 \\
77 & 99.36632084 & 99.02063911 & 98.83867903 & 98.76968407 & 98.70224432 & 98.61458678 & 95.53780726 \\
78 & 99.45369561 & 99.14760115 & 98.96578246 & 98.90626016 & 98.83570998 & 98.76388736 & 95.58078773 \\
79 & 99.51265238 & 99.24458997 & 99.08723056 & 99.02304262 & 98.96974118 & 98.90300835 & 95.62518203 \\
80 & 99.57231605 & 99.33521656 & 99.19538865 & 99.13685604 & 99.08397875 & 99.02855656 & 95.66844527 \\
81 & 99.62165876 & 99.42442931 & 99.29789141 & 99.24515551 & 99.18902641 & 99.13812848 & 95.7151017 \\
82 & 99.66350817 & 99.50841088 & 99.39318364 & 99.34836519 & 99.29845694 & 99.24048986 & 95.7678376 \\
83 & 99.70196438 & 99.57387127 & 99.47504446 & 99.44521263 & 99.3934664 & 99.33762007 & 95.82637021 \\
84 & 99.73603771 & 99.63212111 & 99.5443222 & 99.52198366 & 99.48253777 & 99.43163985 & 95.87627846 \\
85 & 99.76657646 & 99.68853298 & 99.60384449 & 99.5870199 & 99.5552087 & 99.51378344 & 95.94216299 \\
86 & 99.79541862 & 99.73688601 & 99.66605306 & 99.64936986 & 99.62250706 & 99.58489915 & 96.01186487 \\
87 & 99.82624014 & 99.78226999 & 99.72175801 & 99.70196438 & 99.68556393 & 99.65219752 & 96.09061527 \\
88 & 99.85494092 & 99.82496769 & 99.78113893 & 99.75682103 & 99.74579315 & 99.71779929 & 96.23001903 \\
89 & 99.88109679 & 99.8625756 & 99.82440216 & 99.80121533 & 99.79994288 & 99.77901818 & 96.40575825 \\
90 & 99.91220108 & 99.8960834 & 99.86059624 & 99.84363027 & 99.83910601 & 99.82553323 & 96.58036641 \\
91 & 99.93821558 & 99.92280481 & 99.89495234 & 99.88307616 & 99.87586562 & 99.8676654 & 96.84560115 \\
92 & 99.95687815 & 99.94486058 & 99.9246428 & 99.91573566 & 99.90965618 & 99.90739405 & 97.1023529 \\
93 & 99.97299582 & 99.9639473 & 99.94952623 & 99.94585026 & 99.93821558 & 99.93863972 & 97.46160741 \\
94 & 99.98260988 & 99.97638902 & 99.97243029 & 99.9673405 & 99.96324039 & 99.96055411 & 97.90342967 \\
95 & 99.99052733 & 99.98798243 & 99.98289264 & 99.98218573 & 99.9807719 & 99.97808562 & 98.49596634 \\
96 & 99.99462744 & 99.99547574 & 99.99434468 & 99.99363776 & 99.99250669 & 99.99123425 & 99.95786783 \\
97 & 99.99646542 & 99.99872755 & 99.9983034 & 99.9983034 & 99.99802064 & 99.99787925 & 99.96620943 \\
98 & 99.99773787 & 99.99985862 & 99.99971723 & 99.99971723 & 99.99985862 & 99.99957585 & 99.97511657 \\
99 & 100 & 100 & 100 & 100 & 100 & 100 & 100 \\
100 & 100 & 100 & 100 & 100 & 100 & 100 & 100 \\
}\tenpercent
 
\begin{figure}
    \centering

\begin{tikzpicture}
\begin{axis}[line width=1,enlargelimits=false,label style={font=\bfseries\Large},legend style={at={( 0.6,0.6)}, anchor=north west, font=\small},grid=major, xlabel=\% Error, ylabel=\% Branches]

      \addplot[solid, red] table[x=Error,y=5HL] {\tenpercent};
      \addplot[loosely dotted] table[x=Error,y=Heuristics] {\tenpercent};

    \legend{ML, Heuristics}

\end{axis}

\end{tikzpicture}
    \caption{Prediction Error}
    \label{fig:error_cdf_ten_percent}
\end{figure}
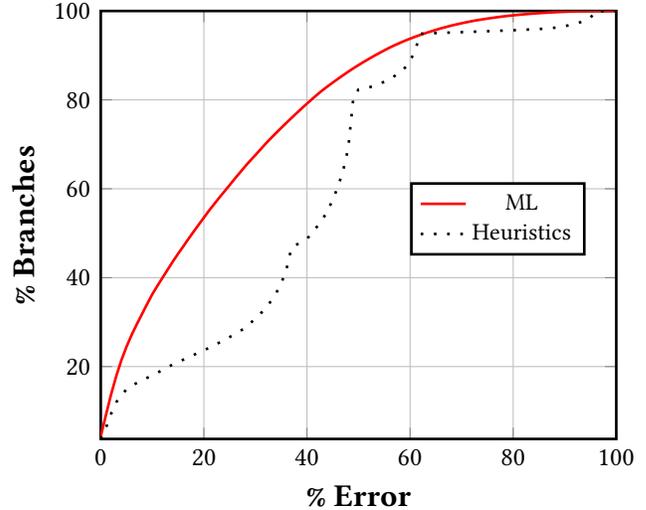 

\pgfplotstableread[row sep=\\,col sep=&]{
Error&ML&Heuristics\\
0 & 2.710644982 & 2.761311243 \\
1 & 7.057810204 & 3.848102549 \\
2 & 12.22830217 & 5.251557988 \\
3 & 16.77559913 & 10.71084765 \\
4 & 20.50970259 & 11.2631099 \\
5 & 24.09687389 & 13.53802503 \\
6 & 27.00511729 & 14.02948776 \\
7 & 29.53589705 & 14.40948472 \\
8 & 31.83107868 & 14.76414855 \\
9 & 34.33399199 & 15.19227846 \\
10 & 36.73557278 & 15.55707554 \\
11 & 38.98008816 & 16.05613822 \\
12 & 40.87753965 & 16.83133202 \\
13 & 42.65085879 & 17.29746162 \\
14 & 44.26457922 & 17.72559153 \\
15 & 45.80990019 & 18.12585499 \\
16 & 47.57308608 & 18.59705122 \\
17 & 49.02467447 & 18.9795815 \\
18 & 50.41799666 & 19.44317779 \\
19 & 51.76571921 & 19.90424077 \\
20 & 53.09824188 & 20.28423773 \\
21 & 54.54476364 & 20.69970107 \\
22 & 55.90008613 & 21.11769773 \\
23 & 57.14900947 & 21.500228 \\
24 & 58.46126564 & 21.93089122 \\
25 & 59.72285555 & 22.40715408 \\
26 & 61.07311142 & 22.86821705 \\
27 & 62.26376856 & 23.27608046 \\
28 & 63.44175913 & 23.7295435 \\
29 & 64.60708314 & 24.30460556 \\
30 & 65.73947408 & 24.97593353 \\
31 & 66.86173177 & 25.48006283 \\
32 & 67.86239043 & 26.18685717 \\
33 & 68.89851548 & 26.97471754 \\
34 & 69.87890764 & 27.86644374 \\
35 & 70.88716624 & 28.89496884 \\
36 & 71.86755839 & 30.59735522 \\
37 & 72.76435122 & 42.94725642 \\
38 & 73.59274459 & 43.33485332 \\
39 & 74.58580331 & 43.80604955 \\
40 & 75.46486295 & 44.22404621 \\
41 & 76.31858945 & 44.84470791 \\
42 & 77.00258398 & 45.60723514 \\
43 & 77.81071085 & 46.39002888 \\
44 & 78.65177079 & 47.27668845 \\
45 & 79.37376501 & 48.43187921 \\
46 & 80.06789279 & 49.96960024 \\
47 & 80.77722045 & 51.4921214 \\
48 & 81.47388154 & 53.75183665 \\
49 & 82.1629427 & 71.88275827 \\
50 & 82.78867102 & 81.55748087 \\
51 & 83.39919947 & 81.70947966 \\
52 & 84.04012768 & 81.87667832 \\
53 & 84.66078938 & 81.9704109 \\
54 & 85.2662512 & 82.16040938 \\
55 & 85.85397983 & 82.35547449 \\
56 & 86.41130871 & 82.56573947 \\
57 & 86.92050464 & 82.84947054 \\
58 & 87.50823327 & 83.1914678 \\
59 & 88.05036226 & 83.59679789 \\
60 & 88.52915843 & 84.21492628 \\
61 & 89.06368749 & 85.34985053 \\
62 & 89.54501697 & 95.18163855 \\
63 & 89.9959467 & 95.22723818 \\
64 & 90.4620763 & 95.25763794 \\
65 & 90.92567259 & 95.27790444 \\
66 & 91.3538025 & 95.31590414 \\
67 & 91.7996656 & 95.33110402 \\
68 & 92.14419618 & 95.35643715 \\
69 & 92.52672645 & 95.37417034 \\
70 & 92.91432335 & 95.39697016 \\
71 & 93.27912043 & 95.4248366 \\
72 & 93.66925065 & 95.45523636 \\
73 & 94.03151441 & 95.47550286 \\
74 & 94.416578 & 95.51096925 \\
75 & 94.74844201 & 95.541369 \\
76 & 95.0955059 & 95.58190201 \\
77 & 95.38937022 & 95.61990171 \\
78 & 95.75923393 & 95.66043472 \\
79 & 96.05816487 & 95.68830116 \\
80 & 96.3165628 & 95.70856766 \\
81 & 96.64336019 & 95.74150073 \\
82 & 96.91949131 & 95.78203374 \\
83 & 97.14495617 & 95.81496681 \\
84 & 97.38562092 & 95.86563307 \\
85 & 97.67188529 & 95.92643259 \\
86 & 97.92014997 & 95.9669656 \\
87 & 98.1785479 & 95.99989867 \\
88 & 98.3913462 & 96.06323149 \\
89 & 98.6041445 & 96.13669757 \\
90 & 98.81947611 & 96.17723058 \\
91 & 99.00440796 & 96.29122967 \\
92 & 99.20707301 & 96.39256219 \\
93 & 99.40213812 & 96.51416122 \\
94 & 99.56933678 & 96.75482596 \\
95 & 99.708669 & 96.98535745 \\
96 & 99.85560116 & 99.97213356 \\
97 & 99.94426711 & 99.97466687 \\
98 & 99.99746669 & 99.9797335 \\
99 & 100 & 100 \\
100 & 100 & 100 \\
}\searchcdf
 
\begin{figure}
    \centering

\begin{tikzpicture}
\begin{axis}[line width=1,enlargelimits=false,label style={font=\bfseries\Large},legend style={at={( 0.6,0.6)}, anchor=north west, font=\small},grid=major, xlabel=\% Error, ylabel=\% Branches]

      \addplot[solid, red] table[x=Error,y=ML] {\searchcdf};
      \addplot[loosely dotted] table[x=Error,y=Heuristics] {\searchcdf};

    \legend{ML, Heuristics}

\end{axis}

\end{tikzpicture}
    \caption{Prediction Error: Search}
    \label{fig:error_cdf_search}
\end{figure}
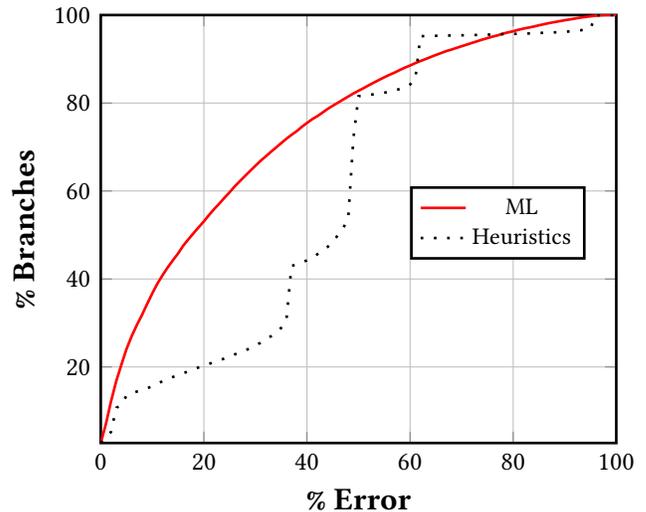 
 
\pgfplotstableread[row sep=\\,col sep=&]{
Error & 0HL & 1HL & 2HL & 3HL & 4HL & 5HL & Heuristics \\
0 & 1.753433489 & 2.806879137 & 3.562854695 & 4.001425142 & 4.206289287 & 4.316143973 & 3.717810598 \\
1 & 3.494990796 & 5.968205763 & 7.656461633 & 8.369880871 & 8.794878538 & 9.097155654 & 6.215060696 \\
2 & 5.452722898 & 9.244901018 & 11.53007643 & 12.66481738 & 13.38035736 & 13.72773569 & 9.371580296 \\
3 & 7.418089688 & 12.13632726 & 14.9649794 & 16.48555489 & 17.13252406 & 17.85386075 & 12.0003167 \\
4 & 9.455703254 & 14.94391332 & 18.22711219 & 19.83647628 & 20.66102265 & 21.43608493 & 13.29241706 \\
5 & 11.26470031 & 17.52514499 & 21.18399882 & 22.70740197 & 23.66513125 & 24.48501198 & 14.96992781 \\
6 & 12.88226462 & 19.79123368 & 23.53604845 & 25.1499368 & 26.15616049 & 27.14810448 & 15.673733 \\
7 & 14.44299291 & 21.89388348 & 25.70062407 & 27.2634731 & 28.45208102 & 29.53535851 & 16.32423674 \\
8 & 15.95550956 & 23.91693459 & 27.71844399 & 29.29458305 & 30.57537276 & 31.81360615 & 16.9152182 \\
9 & 17.3583129 & 25.84681421 & 29.65199958 & 31.33276215 & 32.73782762 & 34.08054314 & 17.47410568 \\
10 & 18.70555268 & 27.68960184 & 31.56321664 & 33.43088769 & 34.88741662 & 36.286544 & 18.00160611 \\
11 & 20.1387534 & 29.51584763 & 33.38479679 & 35.36557434 & 36.76597417 & 38.18206753 & 18.58537703 \\
12 & 21.41643268 & 31.15730569 & 35.00052312 & 37.14530509 & 38.5500878 & 40.02075504 & 19.31392426 \\
13 & 22.73624413 & 32.76935606 & 36.6384466 & 38.88474165 & 40.35060187 & 41.8598667 & 19.9116921 \\
14 & 24.00586457 & 34.47090194 & 38.26590772 & 40.59674988 & 42.11407356 & 43.66872238 & 20.46831746 \\
15 & 25.28043342 & 36.00278808 & 39.90340705 & 42.33986241 & 43.83385786 & 45.38681009 & 20.98295202 \\
16 & 26.60957616 & 37.55644721 & 41.57497971 & 43.99913474 & 45.50048212 & 47.0252991 & 21.54848451 \\
17 & 28.03867677 & 39.17118386 & 43.2370797 & 45.64539982 & 47.10390811 & 48.71765508 & 22.05407056 \\
18 & 29.43525925 & 40.69147658 & 44.88079989 & 47.23963591 & 48.72656221 & 50.33649183 & 22.59669899 \\
19 & 30.8380626 & 42.22746282 & 46.48422588 & 48.83486168 & 50.33635045 & 51.93030378 & 23.16887648 \\
20 & 32.30986091 & 43.80798475 & 48.04170237 & 50.34568174 & 51.92097249 & 53.511674 & 23.69142851 \\
21 & 33.74956525 & 45.40716926 & 49.606248 & 51.87855755 & 53.43872031 & 55.08696476 & 24.1907937 \\
22 & 35.16481031 & 46.97892543 & 51.11791635 & 53.41609901 & 54.93950216 & 56.56173211 & 24.79195473 \\
23 & 36.65965406 & 48.56467854 & 52.60003563 & 54.89100775 & 56.45244296 & 58.02504743 & 25.39467099 \\
24 & 38.2049716 & 50.07069156 & 54.09403109 & 56.38387214 & 57.89624741 & 59.42502312 & 25.95454815 \\
25 & 39.81080111 & 51.54121742 & 55.5540946 & 57.84124937 & 59.32888259 & 60.83574392 & 26.63445959 \\
26 & 41.31752104 & 53.04793736 & 57.10322947 & 59.32336865 & 60.78640121 & 62.27191368 & 27.30701911 \\
27 & 42.87895625 & 54.59891022 & 58.56442405 & 60.73946201 & 62.20772574 & 63.65350955 & 27.97816479 \\
28 & 44.58870236 & 56.12754454 & 59.97952772 & 62.17492485 & 63.63922986 & 65.03821586 & 28.6993601 \\
29 & 46.23708819 & 57.65476504 & 61.48568213 & 63.57659713 & 64.99919412 & 66.343889 & 29.73555701 \\
30 & 47.98161454 & 59.18354074 & 62.88594058 & 64.93373373 & 66.32550919 & 67.59413995 & 30.8380626 \\
31 & 49.78849085 & 60.68601919 & 64.31362735 & 66.31462269 & 67.65253118 & 68.87563658 & 31.92798509 \\
32 & 51.51675814 & 62.14806206 & 65.71289612 & 67.67529386 & 69.017868 & 70.16476789 & 33.22658908 \\
33 & 53.26835365 & 63.60444961 & 67.16221451 & 69.00415384 & 70.2811262 & 71.36949348 & 34.68919748 \\
34 & 55.03267364 & 65.08854825 & 68.5571004 & 70.29088164 & 71.5125732 & 72.54071127 & 36.47288696 \\
35 & 56.83544984 & 66.52952504 & 69.88087058 & 71.53745663 & 72.71701602 & 73.6973666 & 38.6916123 \\
36 & 58.76858128 & 68.00641314 & 71.20775119 & 72.78657652 & 73.89092009 & 74.81853476 & 41.81745177 \\
37 & 60.791491 & 69.5112951 & 72.51568646 & 74.008975 & 75.05690671 & 75.9315027 & 46.74507775 \\
38 & 62.64078224 & 70.99129363 & 73.76636156 & 75.17778928 & 76.21200682 & 77.03457383 & 47.4340377 \\
39 & 64.41443352 & 72.45658831 & 75.01944018 & 76.34306898 & 77.31309858 & 78.10823726 & 48.10179019 \\
40 & 66.26301785 & 73.85119144 & 76.26714624 & 77.52630433 & 78.38619648 & 79.15489652 & 48.77010821 \\
41 & 68.11216771 & 75.22373879 & 77.51683166 & 78.65100707 & 79.47371546 & 80.18458981 & 49.9656439 \\
42 & 70.01447763 & 76.60561743 & 78.76368942 & 79.80893485 & 80.5683036 & 81.20141722 & 51.56921128 \\
43 & 71.95354716 & 78.01351057 & 79.98721897 & 80.92218556 & 81.57169962 & 82.15928788 & 53.10590444 \\
44 & 73.78389307 & 79.40175145 & 81.15433664 & 81.94042681 & 82.51401814 & 83.04165995 & 55.0263114 \\
45 & 75.60391801 & 80.6997899 & 82.24298669 & 82.90126651 & 83.43272567 & 83.88854486 & 57.55353472 \\
46 & 77.30758464 & 81.96842067 & 83.27861806 & 83.83637448 & 84.33446723 & 84.72581571 & 60.4404367 \\
47 & 78.91709011 & 83.15801826 & 84.27296557 & 84.74843701 & 85.19690428 & 85.54032388 & 64.02548855 \\
48 & 80.6160911 & 84.2718345 & 85.22419122 & 85.64183696 & 86.03530619 & 86.33645225 & 69.06523135 \\
49 & 82.17908152 & 85.3491739 & 86.14615056 & 86.52929882 & 86.81531688 & 87.1050109 & 80.48601862 \\
50 & 83.66049388 & 86.38169484 & 87.0405402 & 87.33687922 & 87.58528937 & 87.8321443 & 82.35793117 \\
51 & 85.05947988 & 87.35921776 & 87.89053553 & 88.14431824 & 88.33009566 & 88.54584631 & 82.59078917 \\
52 & 86.38240176 & 88.33589237 & 88.72158553 & 88.90806987 & 89.05143235 & 89.21981965 & 82.84598571 \\
53 & 87.70998928 & 89.24145127 & 89.51036197 & 89.62997209 & 89.73360592 & 89.88474448 & 83.05396028 \\
54 & 89.00251379 & 90.09300182 & 90.28528287 & 90.33646356 & 90.40192394 & 90.5161615 & 83.57396741 \\
55 & 90.10261587 & 90.89266476 & 91.01920266 & 91.02853394 & 91.05341737 & 91.13683341 & 84.09128825 \\
56 & 91.15775812 & 91.61230486 & 91.71028336 & 91.68879313 & 91.6822895 & 91.73898413 & 84.68057311 \\
57 & 92.11265973 & 92.28387469 & 92.34721433 & 92.29023693 & 92.24909444 & 92.29433704 & 85.37221935 \\
58 & 92.99149722 & 92.90638458 & 92.90836394 & 92.85166931 & 92.77447413 & 92.80854746 & 86.28611985 \\
59 & 93.82141615 & 93.50627317 & 93.43925757 & 93.37422133 & 93.30282286 & 93.31144723 & 87.23437646 \\
60 & 94.53271464 & 94.06247437 & 93.94145042 & 93.85464118 & 93.78479792 & 93.77560802 & 88.72328212 \\
61 & 95.18463222 & 94.56961564 & 94.41607356 & 94.31229835 & 94.24174817 & 94.2121991 & 91.2257634 \\
62 & 95.70873946 & 95.05851847 & 94.85987519 & 94.7583621 & 94.67692543 & 94.62136186 & 94.89267607 \\
63 & 96.16272066 & 95.49256466 & 95.28770052 & 95.17063529 & 95.09131936 & 95.01525524 & 94.94314985 \\
64 & 96.59153568 & 95.87359218 & 95.69262178 & 95.570184 & 95.48351614 & 95.38525487 & 94.9854234 \\
65 & 96.96931138 & 96.24401596 & 96.0819909 & 95.94103193 & 95.84630523 & 95.75553727 & 95.03264536 \\
66 & 97.30368246 & 96.58489067 & 96.42272423 & 96.28685505 & 96.20555975 & 96.09966379 & 95.08099839 \\
67 & 97.6384777 & 96.90173025 & 96.74140179 & 96.60468431 & 96.5201372 & 96.40971698 & 95.11719247 \\
68 & 97.9191232 & 97.20923854 & 97.02727846 & 96.89466109 & 96.81534516 & 96.71185271 & 95.15395208 \\
69 & 98.16951271 & 97.49553936 & 97.30113757 & 97.18336543 & 97.0889215 & 96.97991511 & 95.19580149 \\
70 & 98.38370814 & 97.74875654 & 97.56835167 & 97.45991082 & 97.34977336 & 97.23779793 & 95.2409027 \\
71 & 98.59719666 & 97.97779719 & 97.80375457 & 97.68569966 & 97.58687286 & 97.48380456 & 95.28515562 \\
72 & 98.76515981 & 98.1863373 & 98.02728129 & 97.9199715 & 97.80177521 & 97.719773 & 95.32347045 \\
73 & 98.91446038 & 98.36829738 & 98.21404839 & 98.12158383 & 98.02162596 & 97.92152671 & 95.36235081 \\
74 & 99.04877435 & 98.55068161 & 98.39812922 & 98.30368529 & 98.20627232 & 98.10447647 & 95.40321053 \\
75 & 99.16838447 & 98.72316902 & 98.55124714 & 98.47235536 & 98.37720452 & 98.28671932 & 95.44901866 \\
76 & 99.27908746 & 98.88194226 & 98.69913389 & 98.62914924 & 98.54912639 & 98.45906534 & 95.49525094 \\
77 & 99.36632084 & 99.02063911 & 98.83867903 & 98.76968407 & 98.70224432 & 98.61458678 & 95.53780726 \\
78 & 99.45369561 & 99.14760115 & 98.96578246 & 98.90626016 & 98.83570998 & 98.76388736 & 95.58078773 \\
79 & 99.51265238 & 99.24458997 & 99.08723056 & 99.02304262 & 98.96974118 & 98.90300835 & 95.62518203 \\
80 & 99.57231605 & 99.33521656 & 99.19538865 & 99.13685604 & 99.08397875 & 99.02855656 & 95.66844527 \\
81 & 99.62165876 & 99.42442931 & 99.29789141 & 99.24515551 & 99.18902641 & 99.13812848 & 95.7151017 \\
82 & 99.66350817 & 99.50841088 & 99.39318364 & 99.34836519 & 99.29845694 & 99.24048986 & 95.7678376 \\
83 & 99.70196438 & 99.57387127 & 99.47504446 & 99.44521263 & 99.3934664 & 99.33762007 & 95.82637021 \\
84 & 99.73603771 & 99.63212111 & 99.5443222 & 99.52198366 & 99.48253777 & 99.43163985 & 95.87627846 \\
85 & 99.76657646 & 99.68853298 & 99.60384449 & 99.5870199 & 99.5552087 & 99.51378344 & 95.94216299 \\
86 & 99.79541862 & 99.73688601 & 99.66605306 & 99.64936986 & 99.62250706 & 99.58489915 & 96.01186487 \\
87 & 99.82624014 & 99.78226999 & 99.72175801 & 99.70196438 & 99.68556393 & 99.65219752 & 96.09061527 \\
88 & 99.85494092 & 99.82496769 & 99.78113893 & 99.75682103 & 99.74579315 & 99.71779929 & 96.23001903 \\
89 & 99.88109679 & 99.8625756 & 99.82440216 & 99.80121533 & 99.79994288 & 99.77901818 & 96.40575825 \\
90 & 99.91220108 & 99.8960834 & 99.86059624 & 99.84363027 & 99.83910601 & 99.82553323 & 96.58036641 \\
91 & 99.93821558 & 99.92280481 & 99.89495234 & 99.88307616 & 99.87586562 & 99.8676654 & 96.84560115 \\
92 & 99.95687815 & 99.94486058 & 99.9246428 & 99.91573566 & 99.90965618 & 99.90739405 & 97.1023529 \\
93 & 99.97299582 & 99.9639473 & 99.94952623 & 99.94585026 & 99.93821558 & 99.93863972 & 97.46160741 \\
94 & 99.98260988 & 99.97638902 & 99.97243029 & 99.9673405 & 99.96324039 & 99.96055411 & 97.90342967 \\
95 & 99.99052733 & 99.98798243 & 99.98289264 & 99.98218573 & 99.9807719 & 99.97808562 & 98.49596634 \\
96 & 99.99462744 & 99.99547574 & 99.99434468 & 99.99363776 & 99.99250669 & 99.99123425 & 99.95786783 \\
97 & 99.99646542 & 99.99872755 & 99.9983034 & 99.9983034 & 99.99802064 & 99.99787925 & 99.96620943 \\
98 & 99.99773787 & 99.99985862 & 99.99971723 & 99.99971723 & 99.99985862 & 99.99957585 & 99.97511657 \\
99 & 100 & 100 & 100 & 100 & 100 & 100 & 100 \\
100 & 100 & 100 & 100 & 100 & 100 & 100 & 100 \\
}\hlcomparison
 
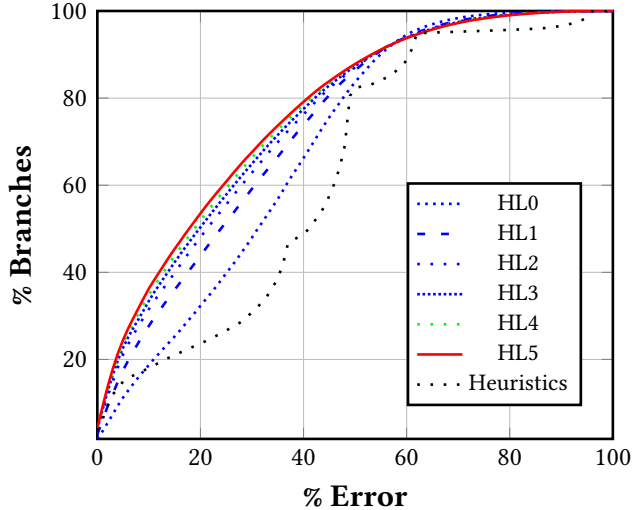
\begin{figure}
    \centering

\begin{tikzpicture}
\begin{axis}[line width=1,enlargelimits=false,label style={font=\bfseries\Large},legend style={at={( 0.6,0.6)}, anchor=north west, font=\small},grid=major, xlabel=\% Error, ylabel=\% Branches]
      \addplot[dotted, blue] table[x=Error,y=0HL] {\hlcomparison};
      \addplot[loosely dashed, blue] table[x=Error,y=1HL] {\hlcomparison};
      \addplot[loosely dotted, blue] table[x=Error,y=2HL] {\hlcomparison};
      \addplot[densely dotted, blue] table[x=Error,y=3HL] {\hlcomparison};
      \addplot[loosely dotted, green] table[x=Error,y=4HL] {\hlcomparison};
      \addplot[solid, red] table[x=Error,y=5HL] {\hlcomparison};

      \addplot[loosely dotted] table[x=Error,y=Heuristics] {\hlcomparison};

              \legend{HL0, HL1, HL2, HL3, HL4, HL5, Heuristics}

\end{axis}

\end{tikzpicture}
    \caption{Effect of Adding Hidden Layers. HL\textbf{n} refers to a model that uses \textbf{n} hidden layers.}
    \label{fig:error_cdf_ten_percent_hl}
\end{figure} 

\pgfplotstableread[row sep=\\,col sep=&]{
Benchmark & ML & SamplePGO \\
adler32.1 & -1.06 & -12.55 \\
alarmlist & 0.95 & -1.17 \\
bigtable & 5.61 & 6.96 \\
arena\_allocation & 0.76 & 0.12 \\
fiber & 0.43 & -0.27 \\
charset & 1.82 & 2.55 \\
doc\_indexing & -1.94 & -4.09 \\
eigen1 & 0.44 & -0.57 \\
eigen2 & 0.20 & 0.28 \\
eigen3 & 0.43 & -1.61 \\
entropy\_coding & -1.49 & -5.90 \\
fingerprint1 & 0.49 & 1.03 \\
fingerprint2 & 0.11 & 0.00 \\
adler32.2 & 1.85 & 0.07 \\
gipfeli & -4.73 & -2.85 \\
storage1 & 0.56 & 3.74 \\
ocr1 & -3.92 & -1.45 \\
ocr2 & -0.68 & 1.25 \\
reed\_solomon & -0.50 & -0.07 \\
sstable1 & 5.67 & 1.89 \\
plaque\_bench & 2.23 & 5.21 \\
protobuf1 & 1.54 & 2.29 \\
protobuf2 & 0.61 & 1.41 \\
protobuf3 & 5.99 & 2.71 \\
protobuf4 & -6.43 & 0.18 \\
protobuf5 & 1.58 & -0.07 \\
protobuf6 & 8.10 & 4.45 \\
protobuf7 & 1.56 & 0.25 \\
speech\_avx & -0.21 & 0.81 \\
speech\_sse & 0.02 & -0.04 \\
sstable2 & -0.39 & -3.18 \\
hashtable1 & -0.28 & -0.71 \\
hashtable2 & 6.28 & 0.65 \\
hashtable3 & 5.39 & 0.42 \\
event\_manager & 5.96 & -9.26 \\
tcmalloc & 0.34 & 2.15 \\
monitoring & -0.22 & -0.43 \\
vision & 4.77 & 17.27 \\
snappy1 & 0.53 & 2.33 \\
snappy2 & -0.83 & 0.02 \\
Geomean & 1.0 & 0.25 \\
}\benchmarkstats
 
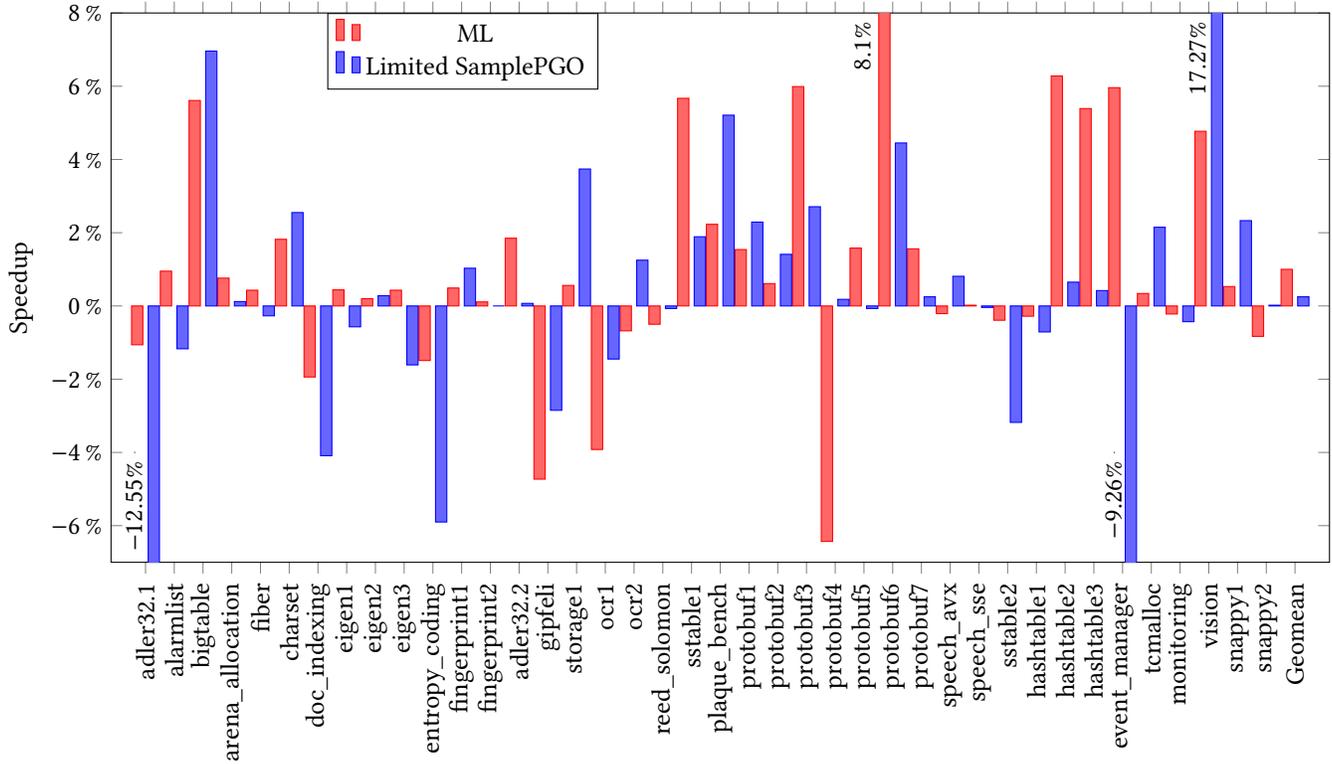
\begin{figure*}
    \centering

\begin{tikzpicture}
    \begin{axis}[
            ybar,
            bar width=.15cm,
            enlarge x limits=0.03,
            width=\textwidth,
            height=.5\textwidth,
            ymin=-7,
            ymax=8,
            ylabel=Speedup,
            legend style={at={(0.4,1)},
                legend columns=1},
            symbolic x coords={adler32.1, alarmlist, bigtable, arena\_allocation, fiber, charset, doc\_indexing, eigen1, eigen2, eigen3, entropy\_coding, fingerprint1, fingerprint2, adler32.2, gipfeli, storage1, ocr1, ocr2, reed\_solomon, sstable1, plaque\_bench, protobuf1, protobuf2, protobuf3, protobuf4, protobuf5, protobuf6, protobuf7, speech\_avx, speech\_sse, sstable2, hashtable1, hashtable2, hashtable3, event\_manager, tcmalloc, monitoring, vision, snappy1, snappy2, Geomean},
            xtick=data,
            xticklabel style={rotate=90},
            scaled ticks=false,
            yticklabel=\pgfmathparse{\tick}\pgfmathprintnumber{\pgfmathresult}\,\%,
            yticklabel style={/pgf/number format/.cd,fixed,precision=2},
            %restrict y to domain*=-7:8, % Cut values off at 14            
            %visualization depends on=rawy\as\rawy,
            %after end axis/.code={ % Draw line indicating break
            %\draw [ultra thick, white, decoration={snake, amplitude=1pt}, decorate] (rel axis cs:0,1.05) -- (rel axis cs:1,1.05);
            %},
            %nodes near coords,
            %ymin=-15,ymax=15
        ]
        \addplot[red,fill=red!60!white]  table[x=Benchmark,y=ML]{\benchmarkstats};
            \draw[xshift=-0.4cm]  (axis cs:protobuf6, 8) -- (axis cs:protobuf6, 9)
      node [pos=0, rotate=90, anchor=east] {\(8.1\%\)};
        %\draw [->, thick, white, xshift=-0.2cm] (protobuf6, 6) -- (protobuf6, 8)
      %node [pos=0, rotate=90, anchor=east] {\(8\decimalpoint1\)};
        \addplot[blue,fill=blue!60!white]  table[x=Benchmark,y=SamplePGO]{\benchmarkstats};
                    \draw[xshift=-0.15cm]  (axis cs:vision, 8) -- (axis cs:vision, 9) node [pos=0, rotate=90, anchor=east] {\(17.27\%\)};
        \draw[xshift=-0.5cm]  (axis cs:tcmalloc, -4) -- (axis cs:tcmalloc, -4) node [pos=0, rotate=90, anchor=east] {\(-9.26\%\)};
        \draw[xshift=-0.15cm]  (axis cs:adler32.1, -4) -- (axis cs:adler32.1, -4) node [pos=0, rotate=90, anchor=east] {\(-12.55\%\)};
        \legend{ML, Limited SamplePGO}
    \end{axis}
\end{tikzpicture}
    \caption{Benchmark Suite Results}
    \label{fig:benchmark_results}
\end{figure*}    

We  now present an evaluation of our ML system for branch probability estimation. Figure~\ref{fig:agg_metrics_ten_percent} compares the ML based prediction with current compiler heuristics using a set of aggregate metrics. For this experiment, we set aside 10\% of all our examples randomly and trained the model with the remaining 90\% of examples. The metrics presented are from the 10\% that was not used for training. The metrics include root mean-squared error, mean absolute error and cross entropy. We see that ML results in 20-50\% accuracy improvement depending on the metric used.
 We also present a \emph{closeness} metric. From the figure, we see that the closeness value for ML is 0.67. This means for 67\% of branches, the ML-predicted value is closer to the actual probability than the heuristic predicted value. Unlike the other three metrics, a larger value of closeness is better.

Figure~\ref{fig:agg_metrics_search} shows the same four metrics from a different dataset. In this case, the evaluation is done on the branches from \texttt{Search} an important service running in Google datacenters. Again, the model is trained on all branches \emph{excluding} the set of data used for evaluation. The trend is similar to Figure~\ref{fig:agg_metrics_ten_percent}, but the improvements throguh ML decrease a bit.

Figures~\ref{fig:error_cdf_ten_percent} and \ref{fig:error_cdf_search} present a more detailed picture of the accuracy of the ML estimation relative to compiler heuristics. This plots the cumulative frequency of the branches based on their prediction error. For each branch in our test data, we first compute the absolute difference between the profile-based probability and probability computed by one of the two techniques we evaluate: ML and heuristics. This absolute difference multiplied by 100 is the \emph{prediction error} for a given branch. The  line labeled ''Heuristics'' is the plot for branch probabilities computed by the heuristics in the LLVM compiler. The other line plots the cumulative frequency for ML-based prediction. From these graphs, we can see that using ML results in a larger percentage of branches with smaller errors. For instance, Figures~\ref{fig:error_cdf_ten_percent} and \ref{fig:error_cdf_search} show more than one third of all branches have an error of less than 10\% and more than half of the branches have an error of less than 20\% when ML is used. The heuristics line shows jumps around certain error percentages. This is due to the fact that the heuristics assign a few specific probabilities and a large number of branches are strongly biased. For instance, the heuristics assign 0.5 to many branches and a significant number of these branches have actual probabilities close to 0 or 1 resulting in many of them having an error close to 50\%.

%\begin{figure}
%        \centering
%        \includegraphics[scale=0.3]{hidden_layer_comparison.pdf}
%            \caption{Effect of Adding Hidden Layers}
%
%    \label{fig:error_cdf_ten_percent_hl}
%\end{figure}

Figure~\ref{fig:error_cdf_ten_percent_hl} shows the effect of adding hidden layers to the neural network. The ''HL0'' line is roughly equivalent to using logistic regression. We see clear win as we go from zero hidden layers to one hidden layer. There is again a noticeable difference as we go from one to two hidden layers. As we go from 2 to 5, the improvements are minor. Adding more hidden layers comes with costs. First, the time it takes to train the model increases, which is a one time cost. We noticed a 2.2X increase in training time going from 0 to 5 hidden layers. The other cost is the time for inference, which also increases. In our experience, the inference cost is insignificant relative to the overall compilation time and is not an area of concern. 

%The rest of the lines are computed by an ML model. The line labeled ''HL0'' is the plot for a model that has no hidden layers. Thus the output computed by this model is a \emph{linear} function of its inputs. Even this model clearly outperforms the compiler heuristics using this metric. For instance, 27\% of branches have less than 10\% error using this model, where as the corresponding number of branches for the compiler heuristics is only 19\%. As we add a hidden layer, 35\% of branches fall within the 10\% error range. Adding more layers does increase the accuracy, but we start getting diminishing returns. The model with 3 layers predicts 41\% of branches within the 10\% error range, but adding two more layers only results in 45\% of branches getting within the 10\% error range.

% \begin{figure*}
%    \centering
%    \includegraphics[scale=0.6]{mcd_perflab_results.pdf}
%    \caption{Benchmark Results}
%    \label{fig:benchmark_results}
%\end{figure*}

Figure~\ref{fig:benchmark_results} shows the performance improvements obtained from the ML based branch probability estimation on an internal benchmark suite. This is a set of 40 benchmarks extracted from performance critical applications and libraries. The baseline performance is from binaries compiled with Clang at optimization level \texttt{O3}. The benchmarks are run 25 times and their average is reported here. The blue bar shows the performance improvement (in percentage) of using the ML model in the compilation process. The geomean improvement is 1\%.

To put this number in perspective, we compare the performance using a limited version of SamplePGO. We make two modifications to SamplePGO. First, we create a single profile that combines the profiles of the same set of applications that we use to train the ML model and use this combined profile for all the benchmarks. This ensures that we use the same data for both ML based and SamplePGO versions. This process of combining the profiles require the sample counts to be normalized which might result in some samples being dropped. Second, we discard the function entry counts, call instruction counts, and the global summary from our combined profile. This ensures that both the profile and ML approach provide only branch probabilities. The red bar in Figure~\ref{fig:benchmark_results} shows the performance of this limited SamplePGO. The ML based approach equals or outperforms this limited SamplePGO version in many benchmarks.

The performance of adler32\_1 benchmark shows an important difference between the SamplePGO and ML based approaches. This benchmark computes the adler-32 checksum performance on blocks of varying byte sizes with specialized code sequence based on the block size. The block sizes affect the behavior of branches. Only a few block sizes occur frequently in production and the profile collected by GWP reflects this. SamplePGO optimizes the code based on this behavior. But the input to the \emph{benchmark} version of this code contains many more block sizes. This mismatch causes the SamplePGO optimized binary to perform poorly on some inputs to the benchmark that were rare at profile collection time.
Even though the ML model is trained by the same profile used by SamplePGO, the model assigns probabilities based on input features and tend to do better even on inputs not seen at profile collection.

\pgfplotstableread[row sep=\\,col sep=&]{
    Metric & ML & ML_error & SamplePGO & SamplePGO_error & SPGOML & SPGOML_error\\
    QPS     & 1.21  & 0.201 & 1.83 & 0.158   & 2.13 & 0.212\\
    Latency     & -1.40 & 0.252 & -1.81 & 0.390 & -2.15 & 0.433  \\
    CPU Usage     & -1.42 & 0.248 & -1.01 & 0.299 & -1.64 & 0.291  \\
    }\searchstats
 
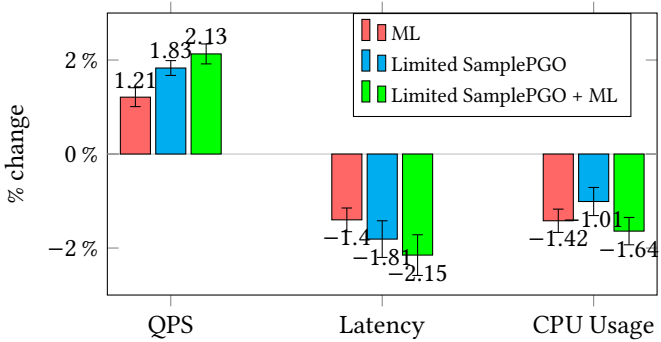
\begin{figure}
    \centering

\begin{tikzpicture}
    \begin{axis}[
            extra y ticks       = 0,
            extra y tick labels = ,
            extra y tick style  = { grid = major },
            ybar,
            bar width=.4cm,
            width=0.5\textwidth,
            height=.3\textwidth,
            legend style={at={(0.7,1)}, anchor=north, font=\footnotesize, legend cell align=left,
                },
            symbolic x coords={QPS, Latency, CPU Usage},
            xtick=data,
            nodes near coords,
            enlarge x limits=0.15,
            ymin=-3,ymax=3,
            ylabel = \% change,
            yticklabel=\pgfmathparse{\tick}\pgfmathprintnumber{\pgfmathresult}\,\%
        ]
        \addplot[style={fill=red!60!white},error bars/.cd, y dir=both, y explicit] table[
        y=ML,
        y error=ML_error]{\searchstats};
        \addplot [style={fill=cyan},
                error bars/.cd,
                y dir=both,
                y explicit
                ] table[x=Metric,y=SamplePGO, y error=SamplePGO_error]{\searchstats};
        \addplot [style={fill=green},
                error bars/.cd,
                y dir=both,
                y explicit
                ] table[x=Metric,y=SPGOML, y error=SPGOML_error]{\searchstats};
        \legend{ML, Limited SamplePGO, Limited SamplePGO + ML}
    \end{axis}
  
\end{tikzpicture}
    \caption{Performance of \texttt{Search} application}
    \label{fig:search_results}
\end{figure}

Figure~\ref{fig:search_results} evaluates the performance of \texttt{Search}. For this experiment, the set of profiles used for training the model excludes the profile collected from running \texttt{Search}. Even though this profile is excluded, many branches in \texttt{Search} have samples as \texttt{Search} shares code with other services. The same set of profiles used for training are combined to generate the profile for the limited SamplePGO experiment. 

The graph shows changes in three metrics -- Queries Per Second (QPS), average latency and CPU usage -- relative to a binary compiled with Clang at -O3. Both the ML model and limited SamplePGO show improvement relative the the baseline with the ML based approach producing 60\% of the improvements of the limited SamplePGO version of the binary. The third column shows the metrics for a version that combines limited SamplePGO and machine learning. In this version, ML is applied to estimate the probability of branches that do not have profile information. This hybrid version outperforms SamplePGO in all the three metrics. This indicates ML has the potential to complement PGO in some cases.
\section{Conclusion}
\label{sec:conclusion}

We have proposed an approach to estimate branch probabilities in the compiler using machine learning. Using various metrics we have demonstrated that our ML based approach significantly out-performs current compiler heuristics for branch probability estimation. On a well-tuned search application, the ML based approach provides 60\% of the performance improvement of a comparable profile guided optimization. The results show that supervised machine learning is a promising approach to approximate profile guided optimization.

%\appendix
%\section{Appendix Title}
%
%This is the text of the appendix, if you need one.

%\acks
%
%Acknowledgments, if needed.

% We recommend abbrvnat bibliography style.

%\bibliographystyle{abbrvnat}
\bibliography{references}

% The bibliography should be embedded for final submission.

\end{document}